\documentclass[sigconf]{acmart}

\usepackage{verbatim}
\usepackage{bm}
\usepackage{bbm}
\usepackage{multirow}
\usepackage{xcolor}
\usepackage{titletoc}
\usepackage{array}
\usepackage{pifont}
\usepackage{nccmath}
\usepackage{bm}
\usepackage{bbm}
\usepackage{titletoc}
\usepackage{subcaption}
\usepackage{cleveref}

\usepackage{physics}
\usepackage{amsmath}
\usepackage{tikz}
\usepackage{mathdots}
\usepackage{yhmath}
\usepackage{cancel}
\usepackage{color}
\usepackage{siunitx}
\usepackage{array}
\usepackage{multirow}
\usepackage{amssymb}
\usepackage{gensymb}
\usepackage{tabularx}
\usepackage{extarrows}
\usepackage{booktabs}
\usetikzlibrary{fadings}
\usetikzlibrary{patterns}
\usetikzlibrary{shadows.blur}
\usetikzlibrary{shapes}
\usepackage{algorithm}
\usepackage{algpseudocode}
\usepackage{chngcntr}

\usepackage{tocloft}
\usepackage[titletoc,toc,title]{appendix}

\usepackage{cleveref}
\crefname{table}{Table}{Tables}
\crefname{figure}{Figure}{Figures}
\crefname{algorithm}{Algorithm}{Algorithms}
\crefname{section}{Section}{Sections}
\crefname{appendix}{Appendix}{Appendices}
\crefname{equation}{Eq.}{Eqs.}

\definecolor{soft_red}{RGB}{240, 110, 110}
\definecolor{soft_blue}{RGB}{100, 170, 220}
\definecolor{define_red}{RGB}{255, 20, 20}
\definecolor{define_green}{RGB}{55, 100, 0}

\AtBeginDocument{%
  \providecommand\BibTeX{{%
    \normalfont B\kern-0.5em{\scshape i\kern-0.25em b}\kern-0.8em\TeX}}}

\copyrightyear{2024}
\acmYear{2024}
\setcopyright{acmlicensed}
\acmConference[MM '24] {Proceedings of the 32nd ACM International Conference on Multimedia}{October 28--November 1, 2024}{Melbourne, VIC, Australia.}
\acmBooktitle{Proceedings of the 32nd ACM International Conference on Multimedia (MM '24), October 28--November 1, 2024, Melbourne, VIC, Australia. }
\acmISBN{979-8-4007-0686-8/24/10}
\acmDOI{10.1145/XXXXXX.XXXXXX}

\setcopyright{acmlicensed}
\copyrightyear{2024}
\acmYear{2024}
\acmDOI{XXXXXXX.XXXXXXX}

\begin{document}

\title{Not All Pairs are Equal: Hierarchical Learning for Average-Precision-Oriented Video Retrieval}


\author{Yang Liu}
\affiliation{%
  \institution{SCST, UCAS}
  \city{}
  \country{}
  }
\email{liuyang232@mails.ucas.ac.cn}

\author{Qianqian Xu}
\authornotemark[1]
\affiliation{%
  \institution{IIP, ICT, CAS}
  \city{}
  \country{}
  }
\email{xuqianqian@ict.ac.cn}

\author{Peisong Wen}
\affiliation{%
 \institution{IIP, ICT, CAS}
 \city{}
 \state{}
 \country{}
 }
 \affiliation{%
  \institution{SCST, UCAS}
  \city{}
  \country{}
  }
\email{wenpeisong20z@ict.ac.cn}

\author{Siran Dai}
\affiliation{%
 \institution{IIE, CAS}
 \city{}
 \country{}
}
\affiliation{%
 \institution{SCS, UCAS}
 \city{}
 \country{}
}
\email{daisiran@iie.ac.cn}

\author{Qingming Huang}
\authornotemark[1]
\affiliation{%
 \institution{SCST, UCAS}
 \city{}
 \country{}
}
\affiliation{%
 \institution{IIP, ICT, CAS}
 \city{}
 \country{}
}
\affiliation{%
 \institution{BDKM, CAS}
 \city{}
 \country{}
}
\affiliation{%
 \institution{Peng Cheng Laboratory}
 \city{}
 \country{}
}
\email{qmhuang@ucas.ac.cn}

\renewcommand{\shortauthors}{Yang Liu and Qianqian Xu, et al.}

\begin{abstract}

The rapid growth of online video resources has significantly promoted the development of video retrieval methods. As a standard evaluation metric for video retrieval, Average Precision (AP) assesses the overall rankings of relevant videos at the top list, making the predicted scores a reliable reference for the users. However, recent video retrieval methods utilize pair-wise losses that treat all sample pairs equally, leading to an evident gap between the training objective and evaluation metric.
To effectively bridge this gap, in this work, we aim to address two primary challenges: 
\textbf{a)} The current similarity measure and AP-based loss are suboptimal for video retrieval; 
\textbf{b)} The noticeable noise from frame-to-frame matching introduces ambiguity in estimating the AP loss.
In response to these challenges, we propose the \textit{\textbf{H}ierarchical learning framework for \textbf{A}verage-\textbf{P}recision-oriented \textbf{V}ideo \textbf{R}etrieval (\textbf{HAP-VR})}.
For the former challenge, we develop the TopK-Chamfer Similarity and QuadLinear-AP loss to measure and optimize video-level similarities in terms of AP. For the latter challenge, we suggest constraining the frame-level similarities to achieve an accurate AP loss estimation.
Experimental results present that HAP-VR outperforms existing methods on several benchmark datasets, providing a feasible solution for video retrieval tasks and thus offering potential benefits for the multi-media application.


\end{abstract}

\ccsdesc[300]{Information systems~Video search}
\ccsdesc[300]{Information systems~Learning to rank}

\keywords{Video Retrieval; Average Precision; Hierarchical Similarity Optimization; Self-supervised Learning}



\maketitle
\renewcommand{\thefootnote}{\fnsymbol{footnote}}
\footnotetext[1]{Corresponding authors.}


\begin{figure}[h]
  \centering
  \includegraphics[width=0.98\linewidth]{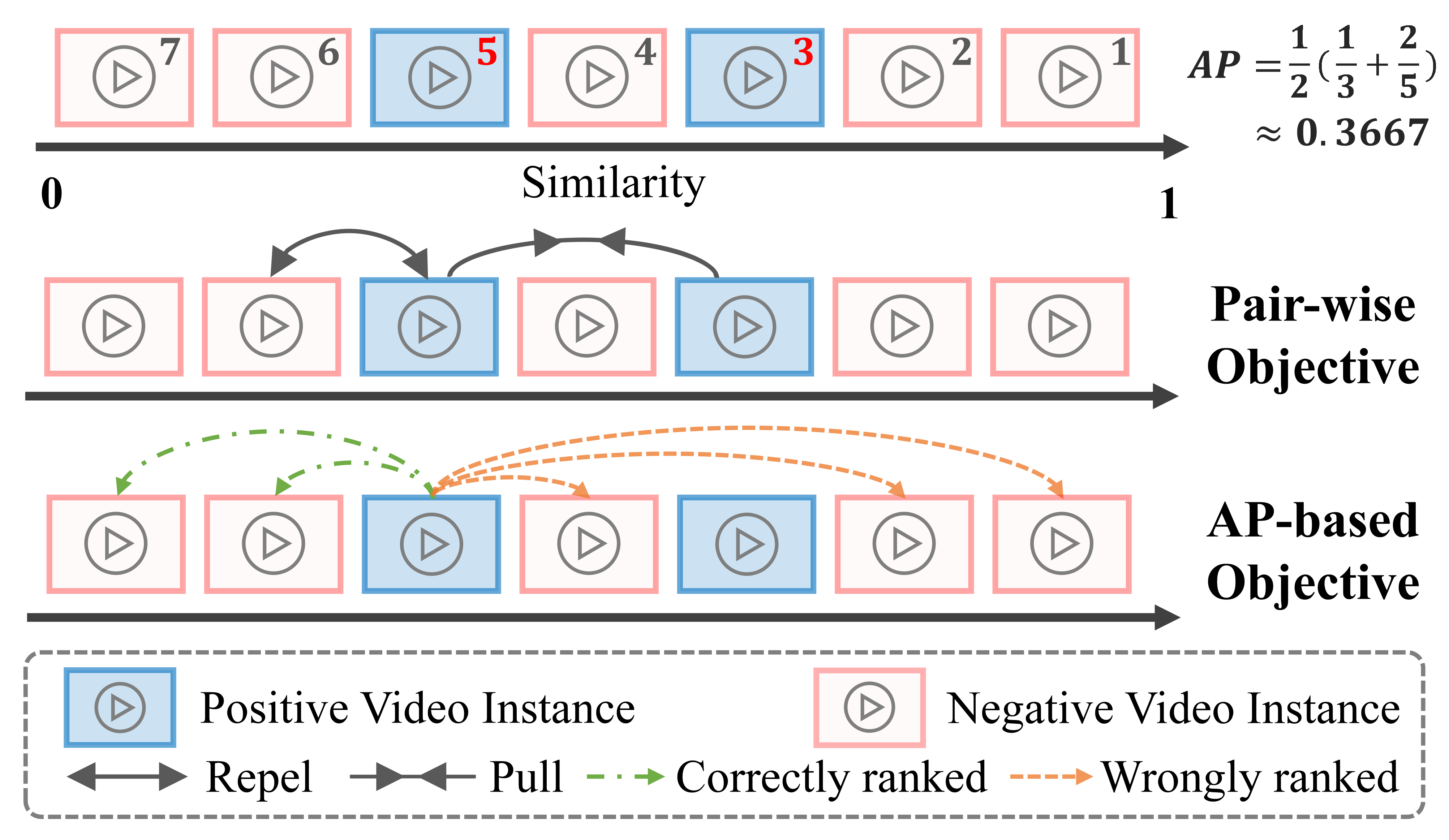}
    \caption{Average Precision (AP) measures the average ranking of positive instances within a list, providing a comprehensive evaluation of the overall performance of the retrieval results. Pair-wise training objectives focus solely on pulling the positive instances closer while repelling the negative ones, failing to align with the AP metric. In contrast, AP-based objectives ensure this alignment by rectifying the rankings of mis-ranked positive-negative pairs in the list.}
   \label{fig:fig1_1}
\end{figure}

\section{Introduction}
\label{sec:introduction}


The rapid expansion of online video resources has made content-based video retrieval a crucial component for multi-media applications such as recommendation, video editing, and online education~\cite{ansari2015content,hu2011survey}. As a fundamental task, video retrieval, aiming to efficiently and effectively rank candidate videos based on their similarities to the query video, has raised a wave of studies in the multi-media community.



Recent video retrieval methods~\cite{kordopatis2017near,shao2021temporal,kordopatis2019visil,kordopatis2022dns,kordopatis2023self} employ neural network models to learn video similarities by aggregating fine-grained embeddings of video frames, which have achieved a remarkable success compared with the early hand-craft approaches~\cite{jiang2014vcdb,douze2010image,tan2009scalable,wang2017compact,chou2015pattern}. Nevertheless, \textbf{these models are typically optimized by pair-wise training objectives such as triplet loss, which are inconsistent with the evaluation metric.} Concretely, as a standard evaluation metric, Average Precision (AP) focuses on the top list by assigning larger weights to the top-ranked positive videos. Since the retrieval results are commonly processed sequentially in downstream tasks, the performance of the top list becomes crucial, thus making AP a more comprehensive metric as it better reflects this practical requirement. However, as shown in \cref{fig:fig1_1}, pair-wise losses treat all mis-ranked video pairs equally, ignoring the relative rankings among the instance list. This leads to an evident gap between the training objective and the evaluation metric, calling for an effective AP-based objective function to bridge this gap.

To solve a similar problem in image retrieval, a promising method is to optimize AP directly. Due to the non-differentiability of AP, existing methods focus on differential approximations for AP~\cite{cakir2019deep,revaud2019learning,cao2007learning,ustinova2016learning,rolinek2020optimizing,brown2020smooth, wen2022exploring, wen2024algorithm}. 
Although these AP optimization methods have achieved notable success in the image field, they cannot be directly applied to videos due to the complexity arising from the additional temporal dimension. Generally, the challenges are two-fold:


\textbf{a) The current similarity measure and the surrogate AP loss are suboptimal for AP-oriented video retrieval.} Typically, in mainstream frameworks, image similarities are measured with the cosine similarity of an embedding pair, while video similarities are aggregated from the redundant temporal-spatial features. Since AP jointly considers the rankings of all instances rather than merely distinguishing whether a pair of videos matches, it is necessary to design a fine-grained similarity measure function for videos. Additionally, existing surrogate AP losses like Smooth-AP~\cite{brown2020smooth} suffer from a vanishing gradient when a sample pair is seriously mis-ranked, leading to inefficient optimization. This phenomenon is more obvious for videos since the various video similarities fall into the gradient vanishing area more frequently.

\begin{figure}
  \centering
  \includegraphics[width=0.98\linewidth]{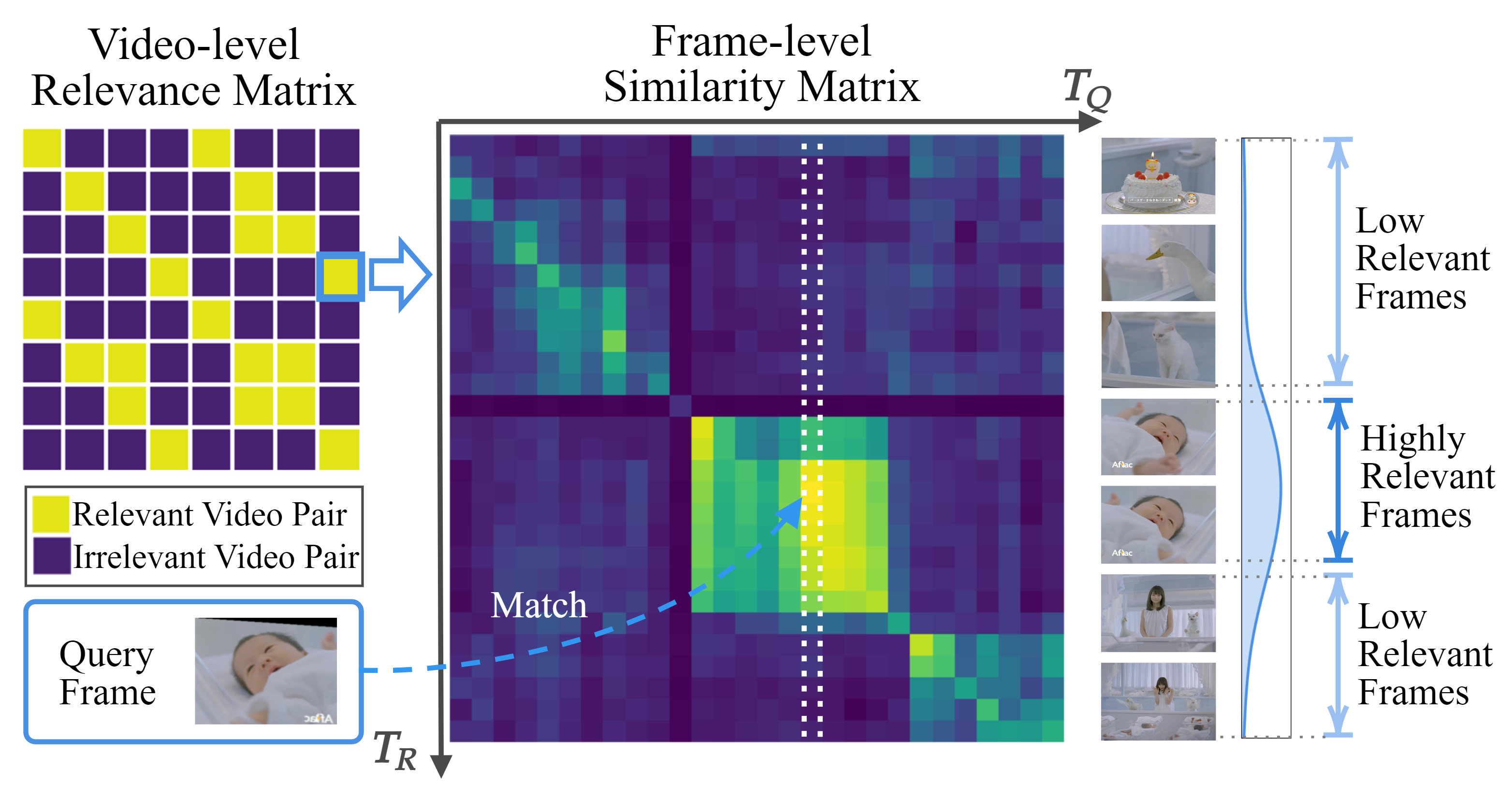}
    \caption{Two relevant videos might not exhibit consistent relevance across all frame pairs due to the obvious redundancy and noise in the temporal dimension. Specifically, only a few consecutive frames in the candidate video are relevant to a given query frame. $\bm{T_Q}$ and $\bm{T_R}$ axes represent the timelines of the query and reference videos, respectively.}
   \label{fig:fig1_2}
\end{figure}

\textbf{b) The noisy frame-to-frame matching leads to a biased AP estimation.} As illustrated in ~\cref{fig:fig1_2}, two relevant videos may not exhibit uniform relevance across all frames. Without fine-grained annotations, this ambiguity leads to false positive matching~\cite{xu2020not}. In this case, the weights of the top-ranked videos might be reduced, hindering the AP loss from concentrating on the top list.

Based on the above considerations, we propose the \textit{\textbf{H}ierarchical learning framework for \textbf{A}verage-\textbf{P}recision-oriented \textbf{V}ideo \textbf{R}etrieval (\textbf{HAP-VR})}, which contains video-level and frame-level constraints as detailed following.

To tackle challenge \textbf{a)}, we propose a topK-based similarity measure and a variant of AP loss with sufficient large gradients.
As a core component of our framework, the proposed \textit{TopK-Chamfer Similarity} aggregates video-level similarities from frame-level similarities. Compared with previous maximum/average aggregations, the TopK-Chamfer Similarity retains fine-grained information while filtering out false correlations, providing refined video similarity for the following AP loss estimation. Another core component is a new surrogate AP loss, namely \textit{QuadLinear-AP}, which enjoys a more reasonable distribution of gradients to rectify mis-ranked positive-negative pairs efficiently.

In search of a solution to challenge \textbf{b)}, we propose to correct the frame-level similarities without requiring fine-grained annotations. Motivated by the recent advance in self-supervised learning~\cite{chen2020simple,he2020momentum,henaff2020data}, we leverage the pre-trained vision model~\cite{caron2021emerging} to extract frame-level representations. Subsequently, we generate pseudo labels indicating the matched frames from the gap between these representations and distill the frame-level information to avoid ambiguity, leading to a more precise estimation of AP loss.

To summarize, the contributions of this work are three-fold:
\begin{itemize}
  \item We develop a self-supervised hierarchical learning framework for Average-Precision-oriented video retrieval, named HAP-VR, to fill the gap between training objectives and evaluation metrics that the previous work has overlooked.
  \item Within HAP-VR, we propose the TopK-Chamfer Similarity and QuadLinear-AP loss to measure and optimize video-level similarities of the AP metric, alongside constraining frame-level similarities to produce a precise estimation of AP loss.
  \item Our experimental evaluation of HAP-VR across several large-scale benchmark datasets often presents a superior performance in terms of AP, ensuring its effectiveness for content-based video retrieval tasks.
\end{itemize}



\begin{figure*}
  \centering
    \includegraphics[width=0.99\linewidth]{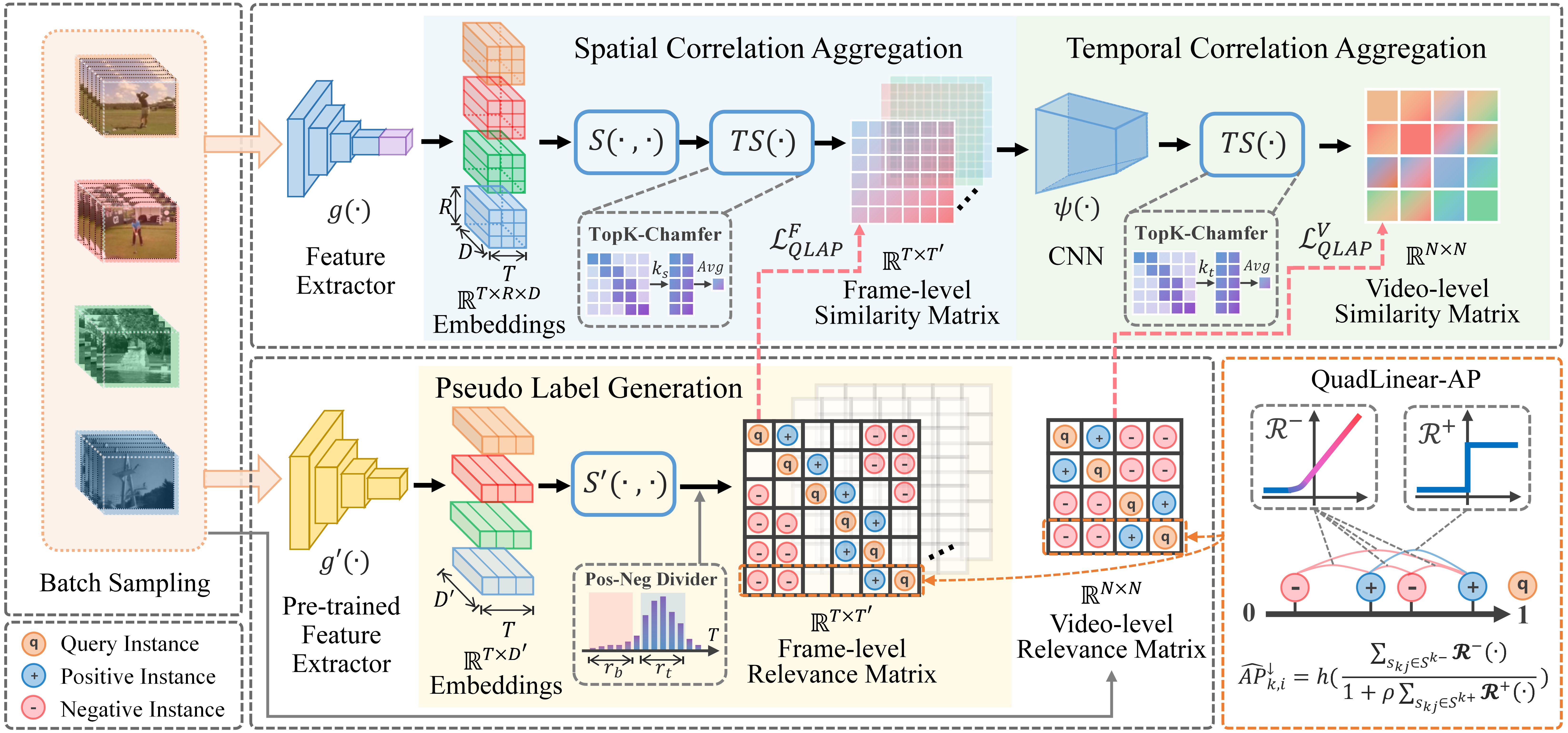}
    \caption{The architecture of our proposed framework. The data batch is processed through a feature extractor to obtain patch-level embeddings. Afterward, we compute frame-level and video-level similarity matrices utilizing spatial and temporal correlation aggregation modules in sequence. Simultaneously, the batch is fed into a pre-trained self-supervised model to generate pseudo labels that indicate frame-level relevance. Ultimately, we apply the QuadLinear-AP to both the frame-level and video-level similarity matrices and backpropagate the loss to optimize the model's parameters.}
    \label{fig:pipeline}
\end{figure*}

\section{Related Work}
\label{sec:related_work}

In this section, we will introduce several previous works that contribute to video retrieval and Average Precision optimization.

\subsection{Video Retrieval}
\label{subsec:related_video}

Based on the granularity of similarity processing, video retrieval methods can be generally classified into two schemes, i.e. coarse-grained method and fine-grained method. 

\subsubsection{Coarse-grained Method}
This kind of method focuses on extracting and aggregating features into a vector space, representing each video by a single vector to compute similarity at the video level. In the early stage, methods such as Bag-of-Words~\cite{shang2010real,cai2011million}, code books~\cite{kordopatis2017near_cnn,liao2018ir} encode videos into a single vector by summarizing the extracted features through statistical aggregation, which neglect the temporal and spatial structures of the video. With the advent of deep learning in the video field, later approaches have started to train deep neural networks with metric learning~\cite{kordopatis2017near,lee2018collaborative}, promoting the transition from coarse-grained methods to fine-grained methods in the subsequent research.

\subsubsection{Fine-grained Method}
This kind of method typically extracts features from frames and thus generates multiple vectors to represent a video. 
Early fine-grained methods focus on designing video temporal alignment solutions, e.g. temporal Hough Voting~\cite{jiang2014vcdb,douze2010image}, graph-based Temporal Network~\cite{tan2009scalable,wang2017compact} and Dynamic Programming~\cite{chou2015pattern}, to match similar segments within the videos through hand-craft algorithms. 
Following the development of methods like TMK~\cite{poullot2015temporal} and LAMV~\cite{baraldi2018lamv}, which use Fourier transform and kernel tricks for spatio-temporal representation learning, there has been a shift towards transformer-based architectures~\cite{shao2021temporal, he2022learn,he2023transvcl,feng2018video,poullot2015temporal}. Among these methods, TCA~\cite{shao2021temporal} adopts a self-attention mechanism to capture temporal relationships among fine-grained features and uses a contrastive learning strategy for training, VRL~\cite{he2022learn} combines CNN with a transformer structure to train a model without labels.

Recent methods concentrate on designing neural networks to learn similarity functions for calculating video-level similarities from original video representations. ViSiL~\cite{kordopatis2019visil} provides a supervised learning method that designs a CNN to obtain video-level similarities from frame-level similarities. Additionally, DnS~\cite{kordopatis2022dns} employs knowledge distillation to train the student networks with ViSiL serving as a teacher network. More recently, S$^2$VS~\cite{kordopatis2023self} proposes a self-supervised learning approach based on an improved structure of ViSiL.
Despite previous studies developing increasingly complex models, their reliance on training with pair-wise objectives has led to a misalignment with the evaluation metric. Furthermore, these efforts typically focus solely on optimizing video-level similarity, neglecting the importance of frame-level similarity learning. In our work, we propose a method that employs an AP-based objective to bridge this gap, hierarchically optimizing AP for both frame-level and video-level similarities during the learning process.

\subsection{Average Precision Optimization}
\label{subsec:AP}

Traditional metric learning methods provide a learning paradigm for retrieval tasks, mapping instances into an embedding space and employing distance metrics to design pair-wise objective functions such as contrastive loss~\cite{chopra2005learning} or triplet loss~\cite{weinberger2009distance}. However, these methods merely focus on increasing the distance between positive and negative instances within pairs or tuples, neglecting to improve the overall ranking of positives more comprehensively. This narrow focus can lead to overfitting, particularly in the face of imbalanced data distribution. A promising method is learning with ranking-based metrics as the target~\cite{wang2022optimizing, wang2022openauc, yang2021learning, yang2022optimizing}, such as AP. 
However, the non-differentiability of ranking terms in AP poses a challenge, obstructing the update of model parameters during backpropagation. Numerous AP optimization methods have been developed to address this issue. 
Listwise approaches~\cite{cao2007learning, cakir2019deep, revaud2019learning, ustinova2016learning} utilize differentiable histogram binning to optimize loss functions based on ranking lists. Others provide structured learning frameworks based on SVM~\cite{yue2007support, mohapatra2014efficient} or conduct direct loss minimization~\cite{hazan2010direct, song2016training} to optimize AP. 
Moreover, BlackBox combinatorial solvers~\cite{rolinek2020optimizing, poganvcic2019differentiation} are proposed to differentiate the ranking terms in AP. More recently, Smooth-AP~\cite{brown2020smooth} introduces the Sigmoid function to approximate the indicator function, offering a simple and efficient way to differentiate AP. However, approximation methods like Smooth-AP neglect the gradient vanishing in the low AP area. To this end, we propose QuadLinear-AP, a novel loss for AP optimization, to designate appropriate gradients to the improperly ranked positive-negative pairs, ensuring the efficiency of the optimization process.

\section{Methodology}
\label{sec:method}

\subsection{Task Definition}
\label{subsec:task_definition}
In the video space $\mathcal{X}$, each video can be seen as a tensor $\bm{V}= \{\bm{v_j} \in \mathbb{R}^{H \times W \times C}\}_{j=1}^{T}$ where $T$, $H$, $W$, and $C$ represent the dimension of time, height, width, and channel, respectively. Given a pair of videos $\bm{V}_1, \bm{V}_2 \in \mathcal{X}$, video similarity learning aims to learn a similarity function $f: \mathcal{X} \times \mathcal{X} \rightarrow \mathbb{R}$, such that $f(\bm{V}_1, \bm{V}_2)$ represents the relevance between $\bm{V}_1, \bm{V}_2$.
During the training stage, at each step, we sample a batch of videos $\bm{B} = \{\bm{V}_{i} \in \mathcal{X}\}_{i=1}^{N}$ where the length of $\bm{V}_i$ is $T_i$. Let $\bm{Y} \in \{0,1\}^{N\times N}$ be the video-level relevance matrix, where $\bm{Y}_{ij} = 1$ if $\bm{V}_i$ and $\bm{V}_j$ are relevant or $\bm{Y}_{ij} = 0$ otherwise.
For clarity, we denote the similarity score as $s_{ij} = f(\bm{V}_i, \bm{V}_j)$, and denote the rankings among positive/negative subsets as $\bm{S}^{k+} = \{s_{ki} = f(\bm{V}_k, \bm{V}_i) | \bm{V}_i \in \bm{B}, \bm{Y}_{ki}=1, k \neq i\}$, $\bm{S}^{k-} = \{s_{ki} = f(\bm{V}_k, \bm{V}_i) | \bm{V}_i \in \bm{B}, \bm{Y}_{ki}=0\}$.


According to the above definition, we aim to optimize $f$ such that $f(\bm{V}_k, \bm{V}_i) > f(\bm{V}_k, \bm{V}_j)$ if $\bm{Y}_{ki} = 1$ and $\bm{Y}_{kj} = 0$, such that it achieves a higher AP score: 
\begin{equation}
\begin{aligned}
\max_{f} ~~ AP(f) &= \frac{1}{N} \sum_{k=1}^N AP_{k}(f), \\
    AP_k(f) &= \frac{1}{|\bm{S}^{k+}|} \sum_{s_{ki}\in \bm{S}^{k+}} \frac {\mathcal{R}(s_{ki}, \bm{S}^{k+})} {\mathcal{R}(s_{ki}, \bm{S}^{k+} \cup \bm{S}^{k-} )},
\label{eq:ap_score}
\end{aligned}
\end{equation}
where $\mathcal{R}(s, \bm{S}) = 1+ \sum_{s' \in \bm{S}} 
\mathcal{H}(s' - s)$ is the descending ranking of $s$ in $\bm{S}$, $\mathcal{H}(\cdot)$ is the Heaviside function~\cite{zhang2020feature}, \textit{i.e.}, $\mathcal{H}(x) = 1$ if $x > 0$ otherwise $\mathcal{H}(x) = 0$.



\subsection{Overview}
\label{subsec:overview}
We aim to design an AP-oriented framework for video similarity learning to align the training objective with the evaluation metric of video retrieval.
As illustrated in \cref{fig:pipeline}, for two videos $\bm{V}, \bm{V}'\in \mathcal{X}$ in a batch, we first utilize a feature extractor $g(\cdot)$ to extract patch-level embeddings $g(\bm{V}), g(\bm{V}') \in \mathbb{R}^{T \times R \times D}$, where $T$, $R$, $D$ are the number of frames, patches, and the embedding dimension, respectively.
Afterward, the patch-to-patch similarities are measured with the cosine similarity, resulting in a patch-level similarity matrix $S(\bm{V}, \bm{V}') \in \mathbb{R}^{T \times R \times R \times T'}$.

Next, we optimize the similarity measure in a hierarchical strategy. At the video level, we aggregate the spatial and temporal correlation to video-level similarities via the proposed TopK-Chamfer Similarity (detailed in ~\cref{subsec:video_oriented}). Following ViSiL~\cite{kordopatis2019visil}, we also apply a CNN to propagate the inter-frame similarities. Afterward, the video-level similarities are input into the proposed QuadLinear-AP loss.
As outlined in ~\cref{subsec:hierarchical}, for the frame-level constraint, we leverage a pre-trained vision model to generate pseudo labels and distill the frame-to-frame similarities to our feature extractor with the QuadLinear-AP loss.




\subsection{Video-oriented AP Optimization}
\label{subsec:video_oriented}
In this subsection, we first implement the similarity function $f$ through a bottom-up video similarity measure to map patch-level embeddings into similarities. Following this, we propose an AP surrogate loss with appropriate gradients for optimization, instructing $f$ to rank the similarities accurately.

\subsubsection{Bottom-up Video Similarity Measure}
\label{subsubsec:bottom-up}

In this subsection, we present the detailed process of feature aggregation. Specifically, given a pair of videos, we first aggregate the patch-level similarities $S(\bm{V}, \bm{V}') \in \mathbb{R}^{T \times R \times R \times T'}$ along the spatial dimension, leading to a frame-level similarity matrix $m_s(\bm{V}, \bm{V}') \in \mathbb{R}^{T\times T'}$. Afterward, we aggregate the temporal dimension as the video-level similarity $f(\bm{V}, \bm{V}') = m_t(\bm{V}, \bm{V}')$. Consider a batch of videos $\bm{B} = \{\bm{V}_{i}\}_{i=1}^{N}$, similarities of all pairs form an $N\times N$ video-level similarity matrix.


Early work utilizes a maximum/average operator to gather the fine-grained features. Kordopatis-Zilos \textit{et al. }~\cite{kordopatis2019visil} suggest that two relevant frames/videos might be similar only in a part of region/period. From this perspective, to gather the spatial features, they propose to focus on the most similar region in $g(\bm{V}')$ for each query patch in $g(\bm{V})$, leading to the Chamfer-Similarity-based aggregation \cite{barrow1977parametric}:
\begin{equation}
    m_s(\bm{V}, \bm{V}')_{x,y} = \frac{1}{R} \sum_{i=1}^R \max_{j=1,\cdots,R} S(\bm{V}, \bm{V}')_{x,i,j,y}.
\end{equation}
The above operator identifies the maximum score for each query patch and averages these scores of all query patches in a frame to reflect the similarity between two frames. A similar operation is performed to gather the temporal features.

However, focusing on the maximum score makes the similarity measure sensitive to spatial noises caused by distractors. Besides, different from the patch-to-patch similarity matrix with a fixed shape, the temporal dimension in videos is flexible and varies greatly. Furthermore, given a query video $\bm{V}_k$ and two relevant candidate videos $\bm{V}_1, \bm{V}_2$, the Chamfer Similarity might assign equal similarities for both $\bm{V}_1$ and $\bm{V}_2$, even if $\bm{V}_2$ contains more relevant frames. Such a phenomenon reduces the distinguishability of positive samples, leading to an ambiguous ranking estimation.

Therefore, we seek a fine-grained similarity measure to estimate a precise AP loss. Specifically, rather than taking the maximum value, we jointly consider the top K scores:
\begin{equation}
    m_s(\bm{V}, \bm{V}')_{x,y} = \frac{1}{RK} \sum_{i=1}^R \sum_{j=1}^K S(\bm{V}, \bm{V}')_{x,i,[j],y},
\end{equation}
where $K=k_s \times R$ and $S(\bm{V}, \bm{V}')_{x,i,[j],y}$ refers to the $j$-th largest value, or formally: $S(\bm{V}, \bm{V}')_{x,i,[1],y}$ $\geq \cdots$ $\geq S(\bm{V}, \bm{V}')_{x,i,[R],y}$.

On top of the frame-level similarities, following ViSiL~\cite{kordopatis2019visil}, we utilize a CNN block $\psi$ to fuse the frame-to-frame similarities:
\begin{equation}
    \bar{m}_s(\bm{V}, \bm{V}') = \psi\left(m_s(\bm{V}, \bm{V}')\right) \in \mathbb{R}^{\frac{T}{s} \times \frac{T}{s}},
\end{equation}
where $s > 1$ is the downsampling factor of $\psi$. In this way, the frame-level similarity is mapped into a learnable measure space. Additionally, it downscales the similarity matrix to reduce the computational burden. Afterward, we utilize the proposed TopK-Chamfer Similarity in the temporal dimension, leading to the video-level similarity:
\begin{equation}
    f(V, V') = m_t(\bm{V}, \bm{V}') = \frac{1}{T} \sum_{i=1}^T \sum_{j=1}^K \bar{m}_s(\bm{V}, \bm{V}')_{i,[j]},
\label{eq:sim_score_fun}
\end{equation}
where $K=k_t \times T'$ and $\bar{m}_s(\bm{V}, \bm{V}')_{i,[1]} \geq \cdots \geq \bar{m}_s(\bm{V}, \bm{V}')_{i,[T']}$. On one hand, compared with the original Chamfer Similarity, the TopK-Chamfer Similarity maintains fine-grained information; on the other hand, compared with the average operator, it avoids the disturbing noise introduced by the irrelevant frames.

\subsubsection{Gradient-Enhanced AP Surrogate Loss}
\label{subsubsec:AP_optimization}
To effectively update the fine-grained similarity measure, in this part, we propose a new surrogate AP loss, such that it enjoys proper gradients in the mis-ranked area.



For a batch of videos $\bm{B} = \{\bm{V}_{i} \in \mathcal{X}\}_{i=1}^{N}$, 
recall that for a query video $\bm{V}_k$, the similarity scores of the relevant and irrelevant videos are denoted as $\bm{S}^{k+} = \{s_{ki} = f(\bm{V}_k, \bm{V}_i) | \bm{V}_i \in \bm{B}, \bm{Y}_{ki}=1, k \neq i\}$ and $\bm{S}^{k-} = \{s_{ki} = f(\bm{V}_k, \bm{V}_i) | \bm{V}_i \in \bm{B}, \bm{Y}_{ki}=0\}$, respectively. 
For the sake of presentation, let $d^k_{ji} = s_{kj} - s_{ki}$.
According to ~\cref{subsec:task_definition}, 
we aim to maximize the AP score, or equivalently minimize the following AP risk of the query video $\bm{V}_k$:
\begin{equation}
\begin{aligned}
    AP^\downarrow_k(f) = 1 - AP_k(f) & = \frac{1}{|\bm{S}^{k+}|} \sum_{s_{ki}\in \bm{S}^{k+}} \frac {
        \sum_{s_{kj}\in \bm{S}^{k-}} \mathcal{H}(d^k_{ji})
    } {
        1 + \sum_{s_{kj}\in \bm{S}^{k+}\cup \bm{S}^{k-}} \mathcal{H}(d^k_{ji})
    }.
\label{eq:ap_risk}
\end{aligned}
\end{equation}

\begin{figure}
  \centering
  \begin{subfigure}{0.47\linewidth}
    \includegraphics[width=1.0\linewidth]{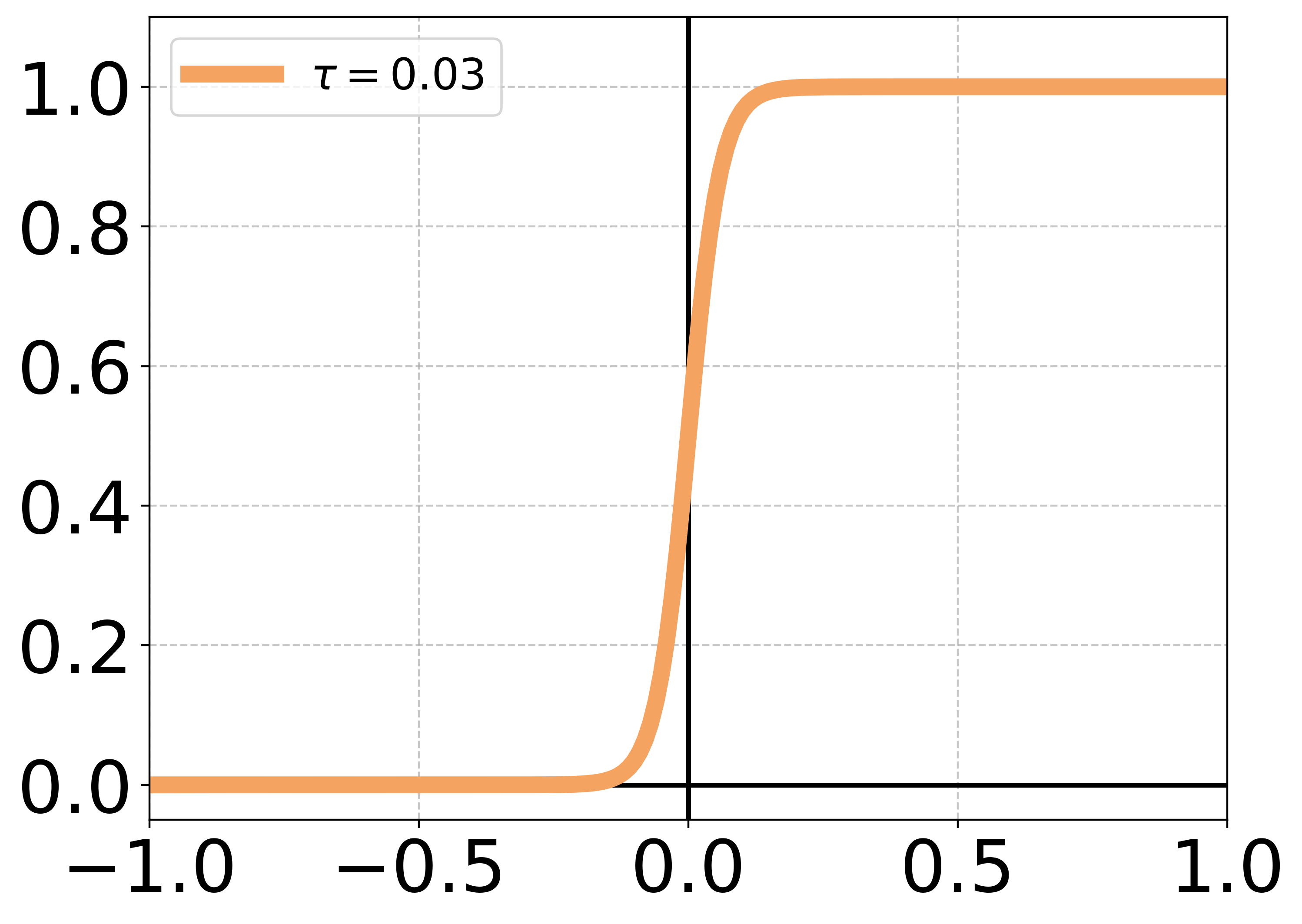}
    \caption{$\mathcal{G}(x;\tau)$ in Smooth-AP.}
    \label{fig:Smooth_AP}
  \end{subfigure}
  \hspace{0.2cm}
  \begin{subfigure}{0.47\linewidth}
    \includegraphics[width=1.0\linewidth]{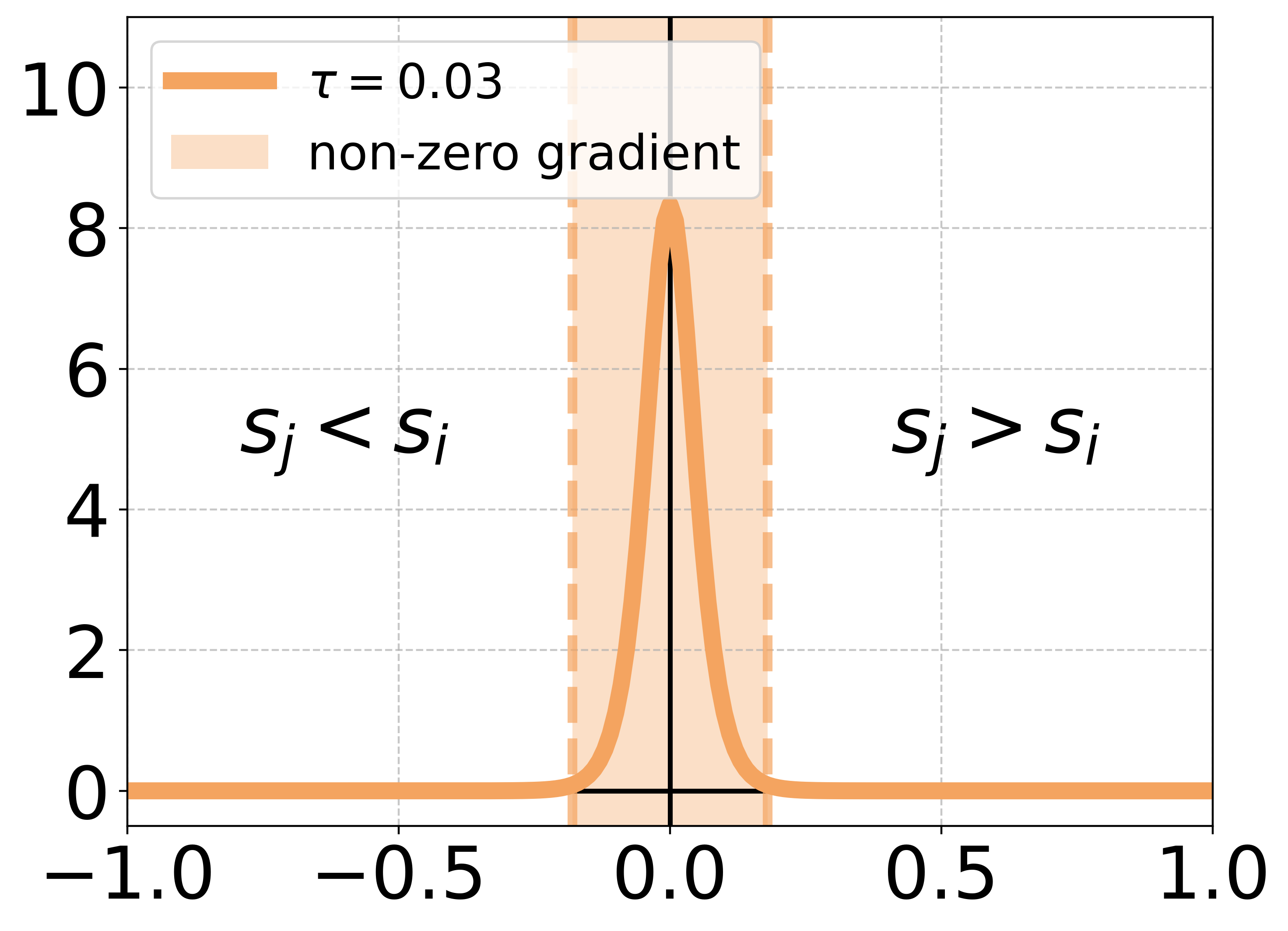}
    \caption{Derivative of $\mathcal{G}(x;\tau)$.}
    \label{fig:Smooth_AP_gradient}
  \end{subfigure}
  
    \begin{subfigure}{0.47\linewidth}
    \includegraphics[width=1.0\linewidth]{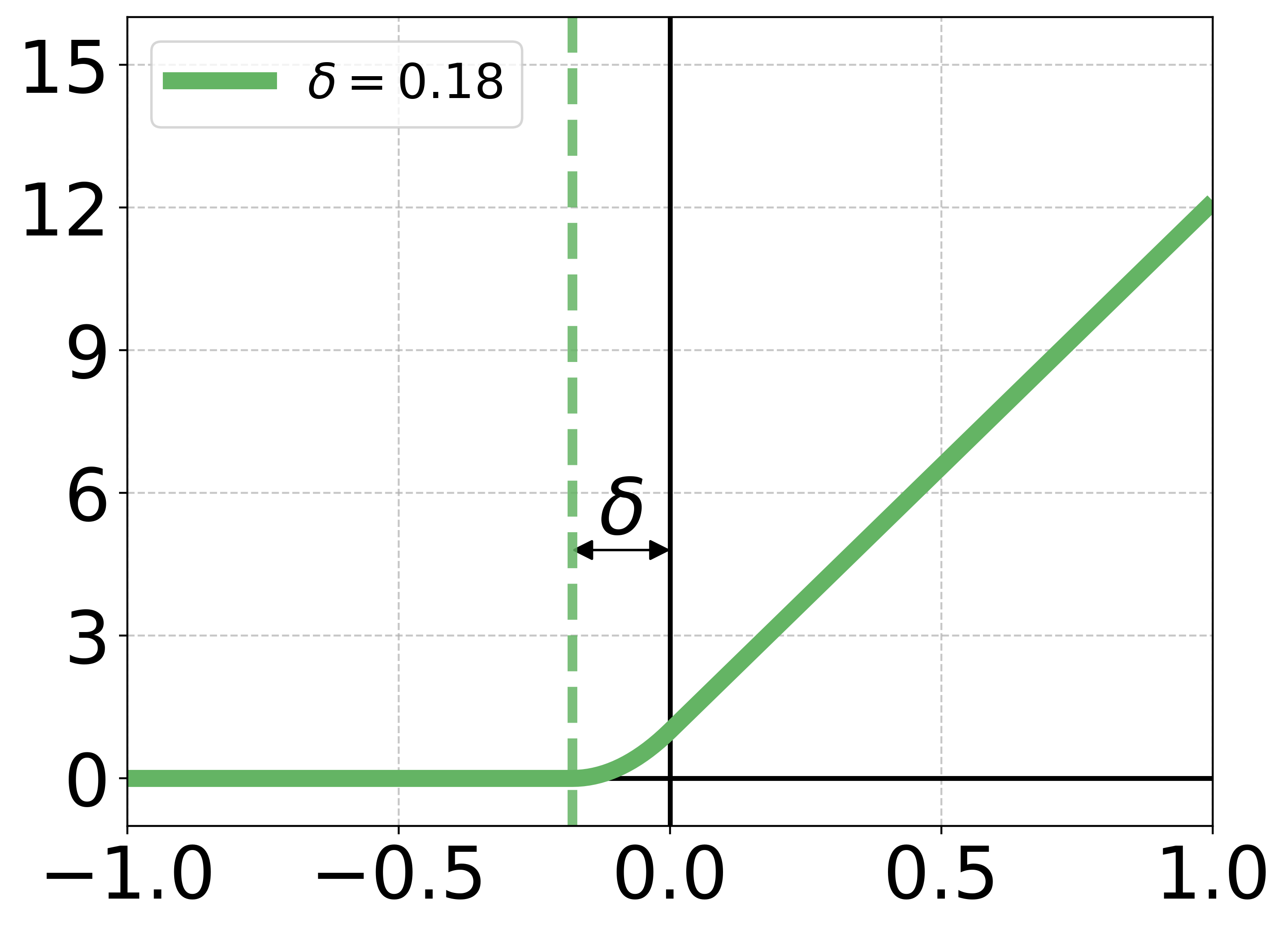}
    \caption{$\mathcal{R^-}(x;\delta)$ in QuadLinear-AP.}
    \label{fig:QuadLinear_AP}
  \end{subfigure}
  \hspace{0.2cm}
    \begin{subfigure}{0.47\linewidth}
    \includegraphics[width=1.0\linewidth]{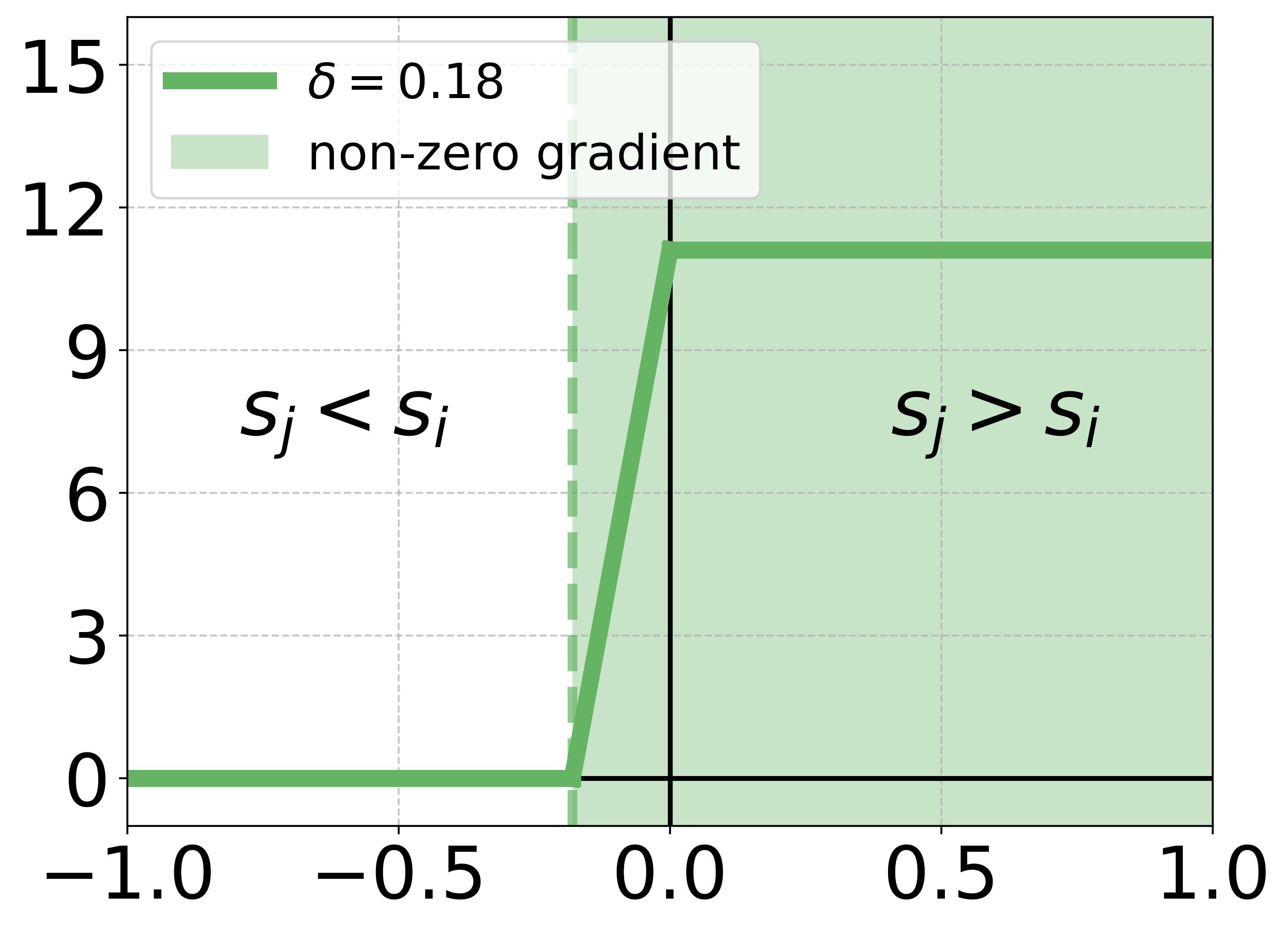}
    \caption{Derivative of $\mathcal{R^-}(x;\delta)$.}
    \label{fig:QuadLinear_AP_gradient}
  \end{subfigure}
  \caption{The curves of Sigmoid function in Smooth-AP and surrogate loss function for positive-negative pairs in QuadLinear-AP and their derivative functions. The colored parts in (b) and (d) represent non-zero gradient areas.}
  \label{fig:Smooth_AP_QuadLinear_AP}
\end{figure}

This AP risk is not differentiable due to the discontinuous function $\mathcal{H}(\cdot)$. To this end, previous methods such as Smooth-AP~\cite{brown2020smooth} employ the Sigmoid function (shown in \cref{fig:Smooth_AP}) as a surrogate:
\begin{equation}
    \mathcal{G}(x;\tau) = {(1+\exp(- x / \tau))}^{-1} \approx \mathcal{H}(x),
\end{equation}
which results in an approximation risk, \textit{i.e.}, $\widetilde{AP}^\downarrow_k(f)$. When $\tau \rightarrow 0$, the $\widetilde{AP}^\downarrow_k(f) \rightarrow AP^\downarrow_k(f)$, thus the approximation error is small.

\textbf{Although Smooth-AP provides a straightforward solution to address the non-differentiable problem of AP, it might suffer from a gradient vanishing issue.} 


Specifically, when the score of a negative instance $s_{kj}$ significantly exceeds that of a positive instance $s_{ki}$, \textit{i.e.} $d^k_{ji} \gg 0$, $d^k_{ji}$ falls into the gradient vanishing area of $\mathcal{G}(x;\tau)$ as depicted in \cref{fig:Smooth_AP_gradient}, \textit{i.e.} $\left. {\frac{d\mathcal{G}(x;\tau)}{dx}} \right|_{x=d^k_{ji}} \approx 0$, leading to slow convergence and sub-optimal solutions. This phenomenon is more evident in video similarity learning since the partial matching property (see ~\cref{subsubsec:bottom-up}) makes $d^k_{ji}$ more likely to fall into the gradient-vanishing area. 

To avoid this issue, we aim to propose a novel AP loss. To begin with, we argue that it is unnecessary to replace all $\mathcal{H}(\cdot)$. Notice that the original AP risk in ~\cref{eq:ap_risk} can be reformulated as:
\begin{equation}
\begin{aligned}
    AP^\downarrow_k(f) = \frac{1}{|\bm{S}^{k+}|} \sum_{s_{ki}\in \bm{S}^{k+}} h\left(\frac {
        \sum_{s_{kj}\in \bm{S}^{k-}} \mathcal{H}(d^k_{ji})
    } {
        1 + \sum_{s_{kj}\in \bm{S}^{k+}} \mathcal{H}(d^k_{ji})
    }\right),
\label{eq:ap_risk_h}
\end{aligned}
\end{equation}
where $h(x) = \frac{x}{1 + x}$ is a monotonically increasing function.
Then, the non-differentiable terms $\mathcal{H}(d^k_{ji})$ can be divided into two types: \textbf{1)} The positive-negative pair ($s_{kj}\in \bm{S}^{k-}$) in the numerator, which should be minimized to ensure the correct ranking; \textbf{2)} The positive-positive pair ($s_{kj}\in \bm{S}^{k+}$) in the denominator, which plays a role of weights.
From this perspective, we only need to ensure that the surrogate loss of the former has an appropriate gradient, while for the latter we can directly use the original rankings such that the importance of each term can be precisely measured.

Motivated by the above observation, for positive-positive pairs we still utilize the Heaviside function:
\begin{equation}
\mathcal{R}^+(x) = \mathcal{H}(x).
\label{eq:R+}
\end{equation}
As for the positive-negative pairs, the derivative of the surrogate loss should be large for the wrongly ranked pairs, \textit{i.e.} $d^k_{ji} + \delta \ge 0$ for a given margin $\delta > 0$. Besides, the surrogate loss should be convex such that the derivative is (non-strictly) monotonically increasing. Therefore, we design the following surrogate loss for positive-negative pairs as displayed in \cref{fig:QuadLinear_AP}:
\begin{equation}
\begin{aligned}
\mathcal{R}^-(x;\delta) =
\left\{
  \begin{array}{lr}
    \mathcal{H}(-x) \cdot \frac{1}{\delta^2}x^2 + \frac{2}{\delta}x+1, & \text{ if } x \ge -\delta . \\
    0, & \text{ if } x < -\delta .
  \end{array}
  \right.
  \label{eq:_QL_R-}
\end{aligned}
\end{equation}

As shown in \cref{fig:QuadLinear_AP_gradient}, $\mathcal{R}^-(x;\delta)$ still provides sufficient large gradients when $d^k_{ji} \gg 0$, and thereby forces the model to optimize $f$ to correct mis-ranked pairs in an effective way. Obviously, the above surrogate loss satisfies our design principles. 
Furthermore, by introducing an extra parameter $\rho$ to adjust the weight of positive-positive pairs, the score distribution between positive and negative instances can be balanced well.
The analysis above induces the formulation of the following AP loss, namely QuadLinear-AP:
\begin{equation}
    \widehat{AP}^\downarrow_k(f) = \frac{1}{|\bm{S}^{k+}|} \sum_{s_{ki}\in \bm{S}^{k+}} h\left(\frac {
        \sum_{s_{kj}\in \bm{S}^{k-}} \mathcal{R}^-(d^k_{ji};\delta)
    } {
        1 + \rho  \sum_{s_{kj}\in \bm{S}^{k+}} \mathcal{R}^+(d^k_{ji})
    }\right),
  \label{eq:eq_QLAP_risk}
\end{equation}
which enjoys the following properties (proved in \cref{subsec:supp_proof}):
\begin{itemize}
  \item \textbf{Differentiable AP optimization} QuadLinear-AP is differentiable for AP term, making it possible to backpropagate gradients in the learning process.
  \item \textbf{Suitable gradients for low AP area.} Persistent and suitable gradients in the loss function force model to correct wrongly ranked positive-negative pairs, avoiding gradient vanishing in the low AP area.
  \item \textbf{Favorable mathematical properties.} QuadLinear-AP is continuous, convex, and (non-strictly) monotonically increasing, ensuring a stable convergence during optimization.
\end{itemize}

As formulated in ~\cref{eq:AP_Loss}, the final AP loss is calculated by averaging QuadLinear-AP across all query videos. Clearly, this objective is aligned with the evaluation metric.
\begin{equation}
\begin{aligned}
\mathcal{L}_{QLAP}^V &= \frac{1}{N} \sum_{k=1}^{N} \widehat{AP}^\downarrow_k(f).
  \label{eq:AP_Loss}
\end{aligned}
\end{equation}

\subsection{Frame Similarity Distillation}
\label{subsec:hierarchical}


As discussed in ~\cref{sec:introduction}, two relevant video instances may not be completely relevant at the frame level due to the noticeable variation in the temporal dimension, \textit{i.e.}, only several frames are highly relevant with a query frame while the others are relatively low in actual. Therefore, solely optimizing $f$ on video-level instances proves inadequate. Next, we dive into the frame-level learning.

Given a query frame, it is hard to locate the relevant frames from another video without fine-grained annotations. A possible route is leveraging self-distillation methods \cite{tarvainen2017mean, caron2021emerging}, which refines image features by distilling ensemble information from a mean teacher to the target model in a self-supervised manner. Unfortunately, since our feature extractor $g$ is trained with video data, it might ignore some image-level information. In this case, the pseudo labels generated by $g$ cannot provide more informative supervision.

Consequently, we introduce another feature extractor $g': \mathcal{X}\mapsto \mathbb{R}^{T \times D'}$, where $D'$ is the embedding dimension. The feature extractor is pre-trained on image data with a self-supervised learning algorithm DINO~\cite{caron2021emerging}, and the parameters are frozen. Given a video pair $\bm{V}, \bm{V}'$, we use $g'$ to extract features for all frames and compute the following frame-level similarities, where $\bm{V}_x$ and $\bm{V}_y'$ are the $x$-th frame of $\bm{V}$ and $y$-th frame of $\bm{V}'$, respectively.
\begin{equation}
    S'(\bm{V}, \bm{V}')_{x,y} = \frac{g'(\bm{V}_x)^\top g'(\bm{V}_y')}{\|g'(\bm{V}_x)\|_2 \|g'(\bm{V}_y')\|_2}.
\end{equation}
As shown in previous study \cite{hamiltonunsupervised}, the similarities are highly correlated to the relevance. However, for different queries, the similarity distributions of its relevant/irrelevant frames are various, hence discretizing them into binary pseudo labels with fixed thresholds is impractical. Instead, we identify the frames with the highest/lowest similarities as positive/negative, leading to the pseudo labels:
\begin{equation}
    \hat{Y}_{x,y} = \left\{
    \begin{array}{cc}
    \begin{aligned}
        1, & \text{ if } S'(\bm{V}, \bm{V}')_{x,y} \geq S'(\bm{V}, \bm{V}')_{x, [r_t \times T']}, \\
        0, & \text{ if } S'(\bm{V}, \bm{V}')_{x,y} \leq S'(\bm{V}, \bm{V}')_{x, [(1 - r_b) \times T']},
    \end{aligned}
    \end{array}
  \right.
\end{equation}
where $r_t, r_b < 1$ are tunable hyperparameters, $S'(\bm{V}, \bm{V}')_{x, [k]}$ refers to the $k$-th largest value in $\{S'(\bm{V}, \bm{V}')_{x,y}\}_{y=1}^{T'}$.


Notice the varying similarity distributions across different video types, it's suboptimal to set a fixed threshold for positive or negative frames to exceed during the training phase. A feasible solution is training the model to learn to rank positive frames above the negative ones. Resembling the method in video-level learning, we optimize the frame-level similarities by $\mathcal{L}_{QLAP}^F$, which can be implemented by substituting the video instances with frame instances.


Following previous methods on ranking optimization \cite{dai2024drauc,shao2024weighted}, we combine a basic loss $\mathcal{L}_{base}$ with the AP losses to promote collaborative optimization between ranking and representation learning. The basic loss comprises the InfoNCE loss~\cite{oord2018representation} to support representation learning and an SSHN loss~\cite{kordopatis2023self} for hard negative mining. 

Ultimately, the total loss for hierarchical similarity learning is formulated in ~\cref{eq:total_loss}, where $\lambda_f$ and $\lambda_v$ are hyperparameters for the trade-off between components.
\begin{equation}
    \mathcal{L} =  \underbrace{\lambda_f\mathcal{L}_{QLAP}^F}_{frame-level} + 
    \underbrace{\lambda_v\mathcal{L}_{QLAP}^V+\mathcal{L}_{base}}_{video-level}
\label{eq:total_loss}
\end{equation}


\section{Experiments}
\label{sec:experiments}

In this section, we begin with a brief overview of the basic settings, including the datasets, evaluation metrics, and implementation details. Next, we compare our proposed learning framework with several previous methods on three benchmark datasets. Finally, we conduct an ablation study to evaluate the performance of different modules. For further details, please see the \cref{sec:supp_detailed_experiments}.


\begin{table*}[htbp]
    \centering
    \small
    \caption{Comparison between video retrieval methods on EVVE, SVD, and FIVR-200K with mAP (\%) of retrieval task and $\bm{\mu}$AP (\%) of detection task. $^\dagger$ indicates the results taken from the original paper. Missing values indicate the lack of implementation or original results. The first and second best results are highlighted in \textbf{\textcolor{soft_red}{soft red}} and \textbf{\textcolor{soft_blue}{soft blue}}, respectively.}
    \resizebox{0.98\textwidth}{!}{%
    \setlength{\extrarowheight}{1pt}
    \begin{tabular}{ccc|ccccc|ccccc}
    \toprule
    \multirow{3}[1]{*}{\textbf{Method}} & \multicolumn{1}{c}{\multirow{3}[1]{*}{\textbf{Label}}}  & \multicolumn{1}{c|}{\multirow{3}[1]{*}{\textbf{Trainset}}} & \multicolumn{5}{c|}{\multirow{1}[0]{*}{\textbf{Retrieval (mAP)}}} & \multicolumn{5}{c}{\multirow{1}[0]{*}{\textbf{Detection ($\bm{\mu}$AP)}}} \\
    \cline{4-13} &  &  & \multicolumn{1}{c}{\multirow{2}[1]{*}{\textbf{EVVE}}} & \multicolumn{1}{c}{\multirow{2}[1]{*}{\textbf{SVD}}} & \multicolumn{3}{c|}{\multirow{1}[1]{*}{\textbf{FIVR-200K}}} & \multicolumn{1}{c}{\multirow{2}[1]{*}{\textbf{EVVE}}} & \multicolumn{1}{c}{\multirow{2}[1]{*}{\textbf{SVD}}} & \multicolumn{3}{c}{\multirow{1}[1]{*}{\textbf{FIVR-200K}}} \\ [1pt] 
    \cline{6-8}\cline{11-13} &  &  &  &  & \multicolumn{1}{c}{DSVR} & \multicolumn{1}{c}{CSVR} & \multicolumn{1}{c|}{ISVR} &  &  & \multicolumn{1}{c}{DSVD} & \multicolumn{1}{c}{CSVD} & \multicolumn{1}{c}{ISVD} \\
    \midrule
        \textbf{DML}$^\dagger$ ~\cite{kordopatis2017near} & \ding{51} & VCDB ($\mathcal{C\&D}$) & 61.10  & 85.00  & 52.80  & 51.40  & 44.00  & 75.50  & /     & 39.00  & 36.50  & 30.00  \\
        \textbf{TMK}$^\dagger$ ~\cite{poullot2015temporal}  & \ding{51} & internal & 61.80  & 86.30  & 52.40  & 50.70  & 42.50  & /  & / & /  & /  & /  \\
        \textbf{TCA}~\cite{shao2021temporal} & \ding{51} & VCDB ($\mathcal{C\&D}$) & 63.08  & \textbf{\textcolor{soft_blue}{89.82}}  & 86.81  & 82.31  & 69.61  & 76.90  & 56.93  & 69.09  & 62.28  & 49.24  \\
        \textbf{ViSiL}$^\dagger$~\cite{kordopatis2019visil} & \ding{51} & VCDB ($\mathcal{C\&D}$) & 65.80  & 88.10  & 89.90  & 85.40  & 72.30  & 79.10  & /     & 75.80  & 69.00  & 53.00  \\
        \textbf{DnS}  ($S_a$)~\cite{kordopatis2022dns} & \ding{51} & DnS-100K & 65.33  & \textbf{\textcolor{soft_red}{90.20}}  & 92.09  & 87.54  & 74.08  & 74.56  & \textbf{\textcolor{soft_red}{72.24}}  & 79.66  & 69.51  & 54.20  \\
        \textbf{DnS} ($S_b$)~\cite{kordopatis2022dns} & \ding{51} & DnS-100K &  64.41  & 89.12  & 90.89  & 86.28  & 72.87  & 75.80  & 66.53  & 78.05  & 68.52  & 53.48  \\
        \cmidrule(lr){1-13}
        \textbf{LAMV}$^\dagger$~\cite{baraldi2018lamv}  & \ding{55} & YFCC100M & 62.00  & 88.00  & 61.90  & 58.70  & 47.90  & 80.60  & /     & 55.40  & 50.00  & 38.80  \\
        \textbf{VRL}$^\dagger$~\cite{he2022learn}  & \ding{55} & internal  & /   & /  & 90.00  & 85.80  & 70.90  & /     & /     & /     & /     & / \\
        \textbf{ViSiL$_f$}$^\dagger$~\cite{kordopatis2019visil} & \ding{55} & ImageNet & 62.70  & /     & 89.00  & 84.80  & 72.10  & 74.60  & /     & 66.90  & 59.50  & 45.90  \\
        \textbf{S$^2$VS}~\cite{kordopatis2023self} & \ding{55} & VCDB ($\mathcal{D}$) & \textbf{\textcolor{soft_blue}{67.17}}  & 88.40  & \textbf{\textcolor{soft_blue}{92.53}}  & \textbf{\textcolor{soft_blue}{87.73}}  & \textbf{\textcolor{soft_blue}{74.51}}  & \textbf{\textcolor{soft_blue}{80.72}}  & 65.04  & \textbf{\textcolor{soft_blue}{86.12}}  & \textbf{\textcolor{soft_blue}{77.41}}  & \textbf{\textcolor{soft_blue}{63.26}}  \\
        \textbf{HAP-VR (Ours)} & \ding{55} & VCDB ($\mathcal{D}$) & \textbf{\textcolor{soft_red}{69.15}} & 89.00 & \textbf{\textcolor{soft_red}{92.83}} & \textbf{\textcolor{soft_red}{88.21}} & \textbf{\textcolor{soft_red}{74.72}} & \textbf{\textcolor{soft_red}{82.88}} & \textbf{\textcolor{soft_blue}{67.87}} & \textbf{\textcolor{soft_red}{88.41}} & \textbf{\textcolor{soft_red}{79.85}} & \textbf{\textcolor{soft_red}{64.79}}  \\
        \bottomrule
      \end{tabular}%
       }

    \label{tab:results_on_methods}%
  \end{table*}%

\subsection{Experimental Setup}

\textbf{Datasets.} Our model is trained on the unlabeled subset of VCDB dataset\cite{jiang2014vcdb} (we denote the core data and distractors as $\mathcal{C}$ and $\mathcal{D}$, respectively) and evaluated on EVVE\cite{revaud2013event}, SVD~\cite{jiang2019svd}, and FIVR-5K/FIVR-200K~\cite{kordopatis2019fivr}. For the FIVR dataset, we report the results of three specific subtasks: DSVR/DSVD, CSVR/CSVD, and ISVR/ISVD.

\textbf{Evaluation Metrics.} For retrieval tasks, we adopt Mean Average Precision (\textbf{mAP}) as the evaluation metric. Specifically, mAP calculates the average AP scores for each query independently and then averages these scores to reflect the model's overall ranking performance. For detection tasks, we employ Micro Average Precision (\textbf{$\bm{\mu}$AP}), a metric widely used in previous studies \cite{law2007video, perronnin2009family, pizzi2022self, kordopatis2023self}. The $\mu$AP calculates the AP across all queries simultaneously, demonstrating the model's capability to consistently apply a uniform threshold across various queries to detect relevant instances.

\textbf{Implementation Details.} Given an input video, we generate two video clips by applying random augmentations that include temporal manipulations~\cite{kordopatis2019visil, kordopatis2023self}, spatial transformations~\cite{cubuk2020randaugment, pizzi2022self}, and other basic augmentations. 
For the feature extractor, we adopt ResNet50~\cite{he2016deep} following \cite{kordopatis2019visil,kordopatis2022dns,kordopatis2023self}, and for the pseudo label generator, we utilize 
DINO~\cite{caron2021emerging} pretrained ViT-small~\cite{dosovitskiy2020image} with a patch size of 16. Our model is trained for 30,000 iterations with a batch size of 64. We use AdamW~\cite{loshchilov2018decoupled} with the Cosine Annealing scheduler for parameters optimization. The learning rate is set to $4 \times 10^{-5}$ with a warm-up period \cite{loshchilov2016sgdr}, and weight decay is set to $1 \times 10^{-2}$. 



\textbf{Competitors.} We evaluate HAP-VR against various leading video retrieval methods, categorized into two types. \textbf{1) Supervised methods} include DML~\cite{kordopatis2017near}, TMK~\cite{poullot2015temporal}, TCA~\cite{shao2021temporal}, ViSiL~\cite{kordopatis2019visil}, DnS~\cite{kordopatis2022dns} with both attention ($S_a$) and binarization ($S_b$) student network. \textbf{2) Unsupervised methods} include LAMV~\cite{baraldi2018lamv}, VRL~\cite{he2022learn}, ViSiL$_f$~\cite{kordopatis2019visil} (untrained ViSiL baseline), and S$^2$VS~\cite{kordopatis2023self}. 



\subsection{Evaluation Results}
\label{subsec:experiment_method}


\begin{figure}
  \centering
  \begin{subfigure}{0.50\linewidth}
    \includegraphics[width=1.0\linewidth]{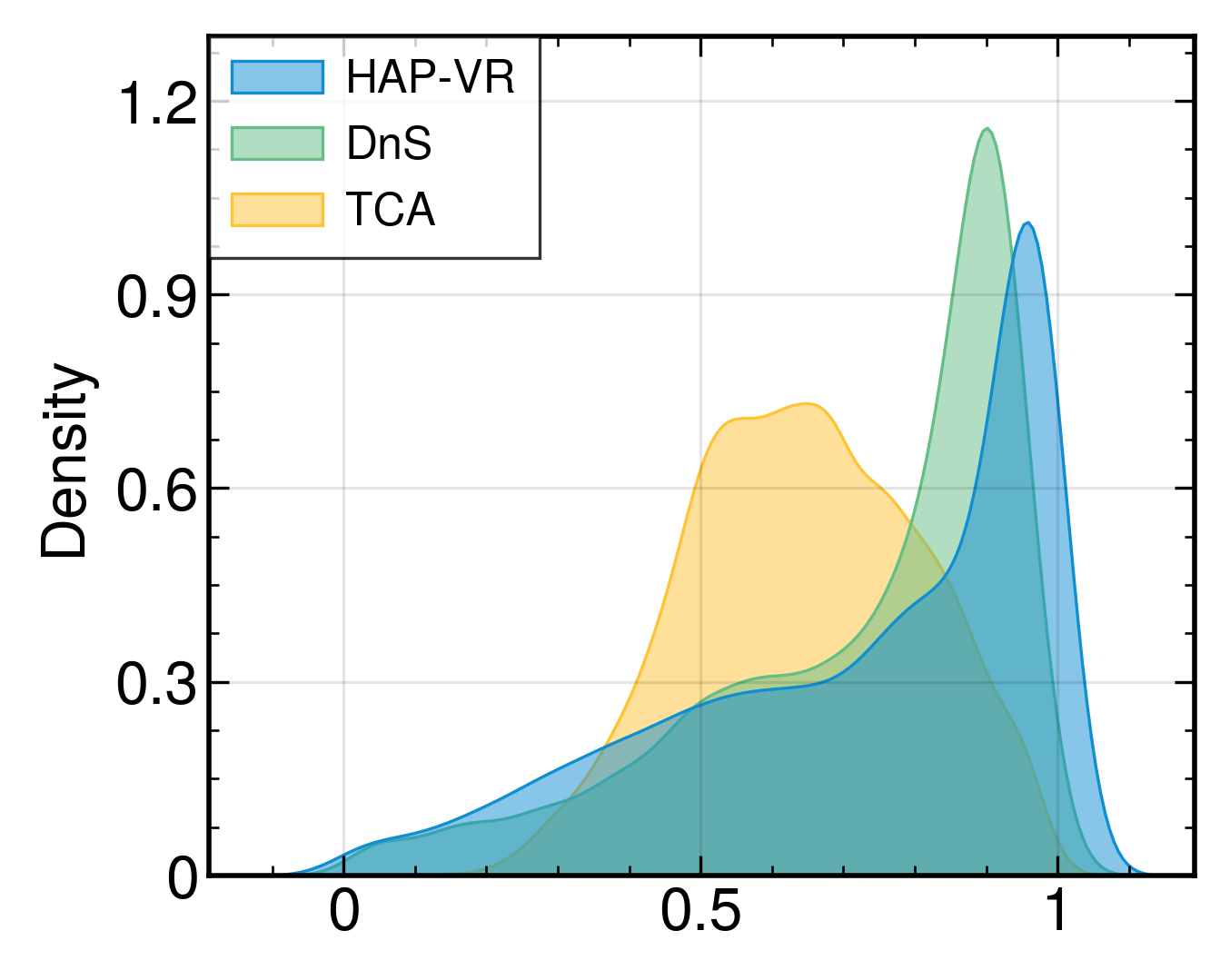}
    \caption{Relevant pair distribution.}
    \label{fig:rele_fivr}
  \end{subfigure}
  \begin{subfigure}{0.47\linewidth}
    \includegraphics[width=1.0\linewidth]{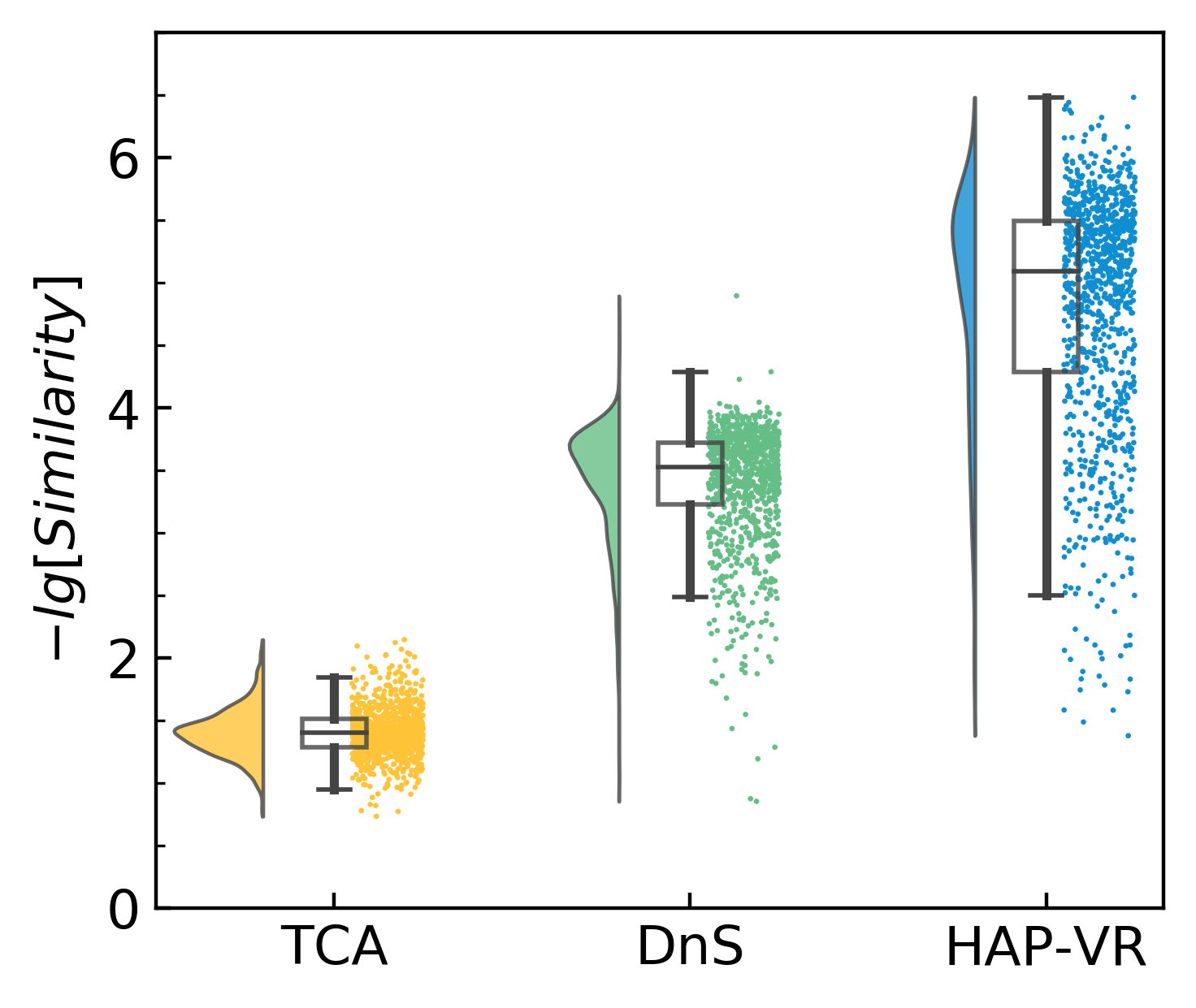}
    \caption{Irrelevant pair distribution.}
    \label{fig:irrele_fivr}
  \end{subfigure}

  \caption{Video similarity distribution of relevant and irrelevant instance pairs for HAP-VR, DnS, and TCA on the DSVD set of FIVR-200K. All similarities are rescaled to $\bm{[0, 1]}$.}
  \label{fig:sim_dist}
\end{figure}


The overall performance on video retrieval and detection tasks above is reported in \cref{tab:results_on_methods}, leading to several key conclusions:
\textbf{1)} HAP-VR stands out among other unsupervised or self-supervised methods in both mAP and $\mu$AP metrics, with an average improvement of \textbf{0.71\%} and \textbf{2.25\%}, respectively. These outcomes underscore the effectiveness of aligning the training objectives with the evaluation metrics, directly enhancing the average precision.
\textbf{2)} Detection tasks enjoy larger performance gains than retrieval tasks. This is primarily due to the more pronounced imbalance between instances in detection tasks. By emphasizing the overall rankings of the positive instances, HAP-VR achieves a more optimal similarity distribution across all queries, resulting in a notable increase in $\mu$AP.
\textbf{3)} Compared with supervised methods, HAP-VR achieves a better overall performance. To investigate the underlying reason, we visualize the similarity distributions in \cref{fig:sim_dist}. Compared with the supervised methods, HAP-VR establishes a clearer margin between scores of relevant and irrelevant pairs. Since annotations are based on the video categories, the supervised model tends to distinguish the pre-defined categories but not video instances. Accordingly, when encountering videos beyond these pre-defined categories, the model is prone to overfitting, which hinders discriminating between negative instances, thereby reducing the model's transferability.




\subsection{Ablation Study}
\label{subsec:ablation_study}
\noindent\textbf{Ablation results on proposed QuadLinear-AP loss.} To validate the effectiveness of the proposed QuadLinear-AP loss, we make a comparison with other commonly used losses, which can be categorized into three types:
\textbf{1) Point-wise losses}, include Mean Absolute Error (MAE) and Mean Squared Error (MSE). These losses measure the discrepancy between predicted scores and actual labels for each item independently.
\textbf{2) Pair-wise losses}, include Contrastive loss~\cite{hadsell2006dimensionality}, Triplet loss~\cite{schroff2015facenet} and Circle loss~\cite{sun2020circle}. These losses focus on distinguishing between the positive and negative instances in pairs.
\textbf{3) List-wise losses}, include FastAP~\cite{cakir2019deep}, DIR~\cite{revaud2019learning}, BlackBox~\cite{poganvcic2019differentiation}, and Smooth-AP~\cite{brown2020smooth}. These approaches optimize the model directly based on ranking metrics such as AP.


\begin{table}[htbp]
  \centering
    \caption{Comparison between QuadLinear-AP and other loss functions on the FIVR-5K with mAP (\%) of retrieval task and $\bm{\mu}$AP (\%) of detection task. The first and second best results are highlighted in \textbf{\textcolor{soft_red}{soft red}} and \textbf{\textcolor{soft_blue}{soft blue}}, respectively.}
    \resizebox{0.49\textwidth}{!}{%
    \setlength{\extrarowheight}{1pt}
    \begin{tabular}{c|ccc|ccc}
    \toprule
    \multirow{2}[2]{*}{\textbf{Losses}} & \multicolumn{3}{c|}{\textbf{Retrieval (mAP)}} & \multicolumn{3}{c}{\textbf{Detection ($\bm{\mu}$AP)}} \\
\cmidrule{2-7} & DSVR  & CSVR  & ISVR  & DSVD  & CSVD  & ISVD \\
    \midrule
    MAE   &  89.07  & 88.03  & 80.86  & 78.08  & 75.69  & 65.26   \\
    MSE   &  89.22  & 88.26  & 80.80  & 78.66  & 76.07  & 65.44 \\
    Contrastive~\cite{hadsell2006dimensionality}  & 88.67  & 88.09  & 80.97  & 75.12  & 74.23  & 67.41   \\
    Triplet~\cite{schroff2015facenet}  & 88.11  & 87.77  & \textbf{\textcolor{soft_blue}{81.21}}  & 72.94  & 73.18  & 69.23  \\
    Circle~\cite{sun2020circle}  & 87.53  & 86.11  & 78.77  & 73.26  & 71.15  & 59.33 \\
    FastAP~\cite{cakir2019deep} & 89.30  & 88.42  & 81.16  & 78.83  & 77.51  & \textbf{\textcolor{soft_blue}{69.95}} \\
    DIR~\cite{revaud2019learning}  & 89.65  & \textbf{\textcolor{soft_blue}{88.57}}  & 80.64  & 78.50  & 76.22  & 65.42 \\
    BlackBox~\cite{poganvcic2019differentiation} & \textbf{\textcolor{soft_blue}{89.70}}  & 88.55  & 80.53  & \textbf{\textcolor{soft_blue}{80.07}}  & 77.37  & 66.00 \\
    Smooth-AP~\cite{brown2020smooth} & 89.36  & 88.33  & 80.73  & 79.85  & \textbf{\textcolor{soft_blue}{77.75}}  & 68.42 \\
    \midrule
    QuadLinear-AP (\textbf{Ours}) &  \textbf{\textcolor{soft_red}{90.80}}  & \textbf{\textcolor{soft_red}{89.68}}  & \textbf{\textcolor{soft_red}{81.31}}  & \textbf{\textcolor{soft_red}{82.92}}  & \textbf{\textcolor{soft_red}{80.03}}  & \textbf{\textcolor{soft_red}{71.45}} \\
    \bottomrule
    \end{tabular}%
    }
  \label{tab:results_on_losses}%
\end{table}%

\begin{figure}
  \centering
  \includegraphics[width=0.98\linewidth]{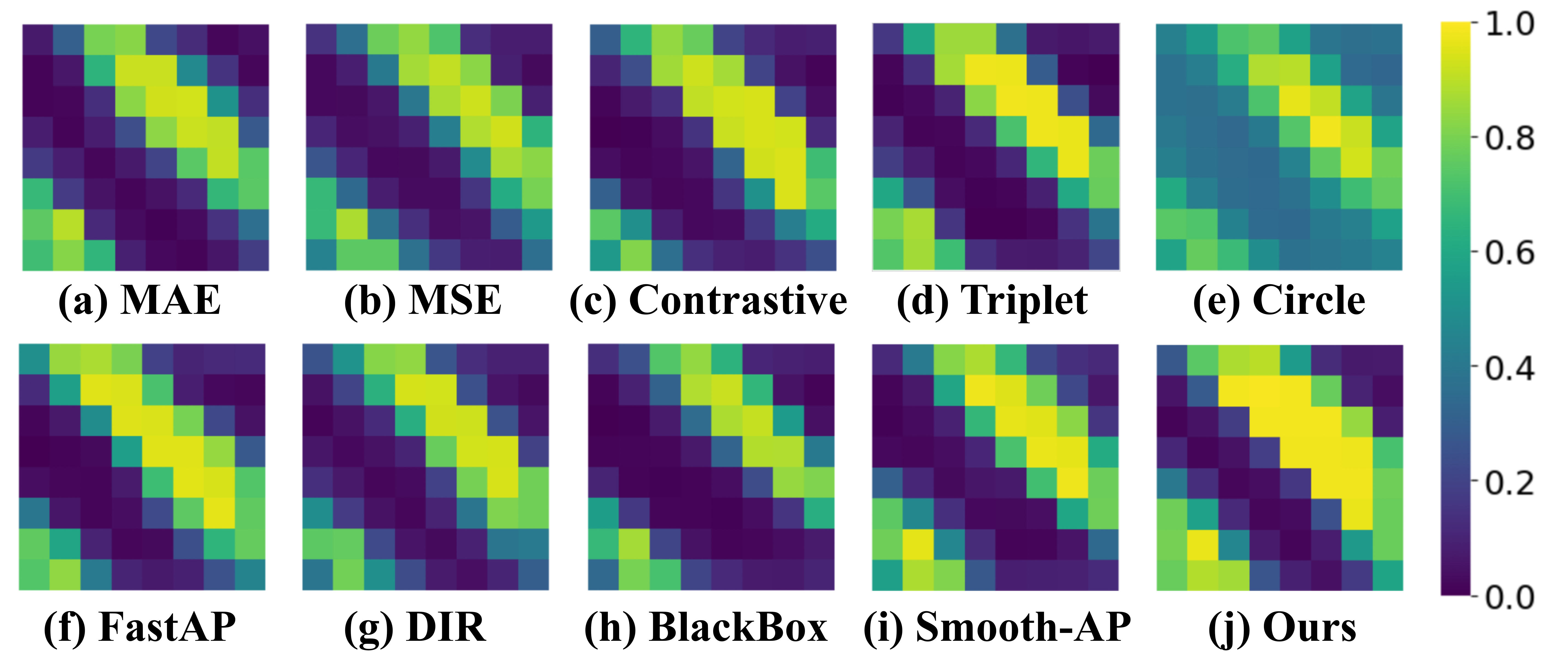}
    \caption{Heatmaps of similarity matrices generated by various losses. In contrast, our QuadLinear-AP distinguishes between relevant and irrelevant instances more clearly.}
   \label{fig:score_matrix_heatmap}
\end{figure}

For a straightforward comparison, we only combine these losses with $\mathcal{L}_{base}$ and train the models using 10\% of the VCDB ($\mathcal{D}$) for 10,000 iterations. Except for the specific hyperparameters associated with each loss, all other settings remain constant to ensure a fair comparison.
The comparison results are presented in \cref{tab:results_on_losses}. From these results, we can draw the following conclusions: \textbf{1)} In general, list-wise losses outperform point-wise and pair-wise losses, supporting our motivation to develop an AP-oriented method for video retrieval tasks. \textbf{2)} QuadLinear-AP achieves an average improvement of about \textbf{1.84\%} on mAP and \textbf{2.87\%} on $\mu$AP over other list-wise losses, reflecting the effectiveness of the proposed AP loss. 
The visualization of frame-level similarity shown in \cref{fig:score_matrix_heatmap} illustrates that QuadLinear-AP presents a clearer distinction between relevant and irrelevant instances compared to other competitors.

\textbf{Ablation results on proposed modules.}
Comparing Line 1 with Line 2 in ~\cref{tab:ablation_module}, the application of the TopK-Chamfer Similarity measure yields average boosts of \textbf{0.59\%} on mAP and \textbf{1.13\%} on $\mu$AP based on the baseline model, demonstrating its efficacy.
Comparing Line 2 with Line 3 shows that incorporating video-level AP optimization enhances retrieval and detection performance with increases of \textbf{0.69\%} and \textbf{2.13\%} respectively. Such improvements reveal the necessity of aligning training objectives with evaluation metrics.
Moreover, implementing a frame-level learning process further improves the overall outcomes, emphasizing the value of learning the internal similarity within the video precisely.

\begin{table}[htbp]
  \centering
    \caption{Results in the ablation study of modules including TopK-Chamfer Similarity measure, video-level AP loss, and frame-level AP loss. Improvements in performance compared to the baseline are denoted with red subscripts.}
    \resizebox{0.47\textwidth}{!}{%
    \begin{tabular}{cccc|cccc}
    \toprule
    \multicolumn{1}{c}{\multirow{2}[0]{*}{\parbox{0.6cm}{$\mathcal{L}_{base}$}}} & \multicolumn{1}{c}{\multirow{2}[0]{*}{\centering \parbox{0.6cm}{ \centering{TopK. \\  Sim.}}}} & \multicolumn{1}{c}{\multirow{2}[0]{*}{\parbox{0.8cm}{$\mathcal{L}_{QLAP}^V$}}} & \multicolumn{1}{c|}{\multirow{2}[0]{*}{\parbox{0.8cm}{$\mathcal{L}_{QLAP}^F$}}} & \multicolumn{1}{c}{\multirow{2}[1]{*}{\textbf{EVVE}}} & \multicolumn{3}{c}{\textbf{FIVR-5K}} \\
    \cline{6-8} &  &  & \multicolumn{1}{c|}{} & \multicolumn{1}{c}{} & \multirow{1}[2]{*}{DSVR/DSVD}  & \multirow{1}[2]{*}{CSVR/CSVD}  & \multirow{1}[2]{*}{ISVR/ISVD} \\ [2pt]
    \midrule
    \multicolumn{8}{c}{\textbf{Retrieval (mAP)}} \\
    \midrule
     \ding{51} &  &  &  & 67.64 & 88.18 & 87.16 & 80.14  \\
     \ding{51} & \ding{51} &  &  & 69.41$_{\textcolor{define_red}{+1.77}}$ & 88.36$_{\textcolor{define_red}{+0.18}}$ & 87.42$_{\textcolor{define_red}{+0.26}}$ & 80.30$_{\textcolor{define_red}{+0.16}}$   \\
     \ding{51} & \ding{51} & \ding{51} &  & 69.55$_{\textcolor{define_red}{+1.91}}$ & 89.39$_{\textcolor{define_red}{+1.21}}$  & 88.54$_{\textcolor{define_red}{+1.38}}$ & 80.79$_{\textcolor{define_red}{+0.65}}$ \\
     \ding{51} & \ding{51} & \ding{51} & \ding{51} & 69.58$_{\textcolor{define_red}{+1.94}}$ & 89.75$_{\textcolor{define_red}{+1.57}}$ & 88.59$_{\textcolor{define_red}{+1.43}}$ & 80.72$_{\textcolor{define_red}{+0.58}}$  \\
    \midrule
    \multicolumn{8}{c}{\textbf{Detection ($\bm{\mu}$AP)}} \\
    \midrule
     \ding{51} &  &  &  & 79.13 & 75.49 & 73.84 & 63.50  \\
     \ding{51} & \ding{51} &  &  & 80.67$_{\textcolor{define_red}{+1.54}}$ & 76.23$_{\textcolor{define_red}{+0.74}}$ & 74.77$_{\textcolor{define_red}{+0.93}}$ & 64.81$_{\textcolor{define_red}{+1.31}}$  \\
     \ding{51} & \ding{51} & \ding{51} &  & 81.09$_{\textcolor{define_red}{+1.96}}$  & 78.64$_{\textcolor{define_red}{+3.15}}$ & 77.04$_{\textcolor{define_red}{+3.20}}$  & 68.23$_{\textcolor{define_red}{+4.73}}$  \\
     \ding{51} & \ding{51} & \ding{51} & \ding{51} & 82.96$_{\textcolor{define_red}{+3.83}}$ & 81.56$_{\textcolor{define_red}{+6.07}}$ & 78.32$_{\textcolor{define_red}{+4.48}}$ & 66.87$_{\textcolor{define_red}{+3.37}}$ \\
    \bottomrule
    \end{tabular}%
    }
  \label{tab:ablation_module}%
\end{table}%

\begin{table}[htbp]
  \centering
    \caption{Results in the ablation study of similarity measure. In particular, $^\ast$ represents using Chamfer Similarity and $^\dagger$ represents using average pooling. The first and second best results are highlighted in \textbf{\textcolor{soft_red}{soft red}} and \textbf{\textcolor{soft_blue}{soft blue}}, respectively.}
    \resizebox{0.47\textwidth}{!}{%
    \begin{tabular}{l|cccc|cccc}
    \toprule
    \multirow{3}[1]{*}{\centering \parbox{0.6cm}{\centering $k_t$}} & \multicolumn{4}{c|}{\multirow{1}[0]{*}{\textbf{Retrieval (mAP)}}} & \multicolumn{4}{c}{\multirow{1}[0]{*}{\textbf{Detection ($\bm{\mu}$AP)}}} \\
    \cline{2-9}  & \multicolumn{1}{c}{\multirow{2}[1]{*}{\textbf{EVVE}}} & \multicolumn{3}{c|}{\multirow{1}[1]{*}{\textbf{FIVR-5K}}} & \multicolumn{1}{c}{\multirow{2}[1]{*}{\textbf{EVVE}}} & \multicolumn{3}{c}{\multirow{1}[1]{*}{\textbf{FIVR-5K}}} \\ [1pt] 
    \cline{3-5}\cline{7-9} &  & \multicolumn{1}{c}{DSVR} & \multicolumn{1}{c}{CSVR} & \multicolumn{1}{c|}{ISVR} &  & \multicolumn{1}{c}{DSVD} & \multicolumn{1}{c}{CSVD} & \multicolumn{1}{c}{ISVD} \\
    \midrule
        0.00$^{\ast}$ & 67.57 & \textbf{\textcolor{soft_blue}{89.52}} & 88.38 & 80.55 & 78.85 & \textbf{\textcolor{soft_blue}{78.78}} & 76.93 & 65.99 \\
        0.03 & 68.98 & \textbf{\textcolor{soft_red}{89.65}} & \textbf{\textcolor{soft_red}{88.69}} & \textbf{\textcolor{soft_red}{80.91}} & 80.70 & \textbf{\textcolor{soft_red}{79.01}} & \textbf{\textcolor{soft_blue}{77.01}} & \textbf{\textcolor{soft_blue}{68.21}} \\
        0.06 & \textbf{\textcolor{soft_red}{69.55}} & 89.39 & \textbf{\textcolor{soft_blue}{88.54}} & \textbf{\textcolor{soft_blue}{80.79}} & \textbf{\textcolor{soft_blue}{81.09}} & 78.64 & \textbf{\textcolor{soft_red}{77.04}} & \textbf{\textcolor{soft_red}{68.23}} \\
        0.10 & \textbf{\textcolor{soft_blue}{69.03}} & 87.87 & 87.12 & 79.69 & 80.75 & 73.19 & 71.75 & 61.96 \\
        0.20 & 68.69 & 85.26 & 85.01 & 78.01 & \textbf{\textcolor{soft_red}{81.54}} & 69.12 & 68.35 & 59.57 \\
        0.30 & 68.27 & 81.54 & 81.98 & 75.93 & 80.04 & 64.30 & 65.41 & 58.41 \\
        1.00$^{\dagger}$ & 55.49 & 61.29 & 64.25 & 62.84 & 77.32 & 37.57 & 44.27 & 42.92 \\
    \bottomrule
    \end{tabular}%
    }
  \label{tab:ablation_top_k_rate}%
\end{table}%

\textbf{Ablation results on similarity measure.}
To validate the effectiveness of the proposed TopK-Chamfer Similarity measure, we evaluate the model varying the top-k rate $k_t$. Note that when $k_t=0.0$, the measure can be seen as the original Chamfer Similarity; when $k_t=1.0$, the measure is equal to average pooling. As indicated in ~\cref{tab:ablation_top_k_rate}, the optimal performance is achieved neither at $k_t=0.0$ nor at $k_t=1.0$. This outcome supports the utility of selecting top-K values. From another perspective, the best performance is obtained when $k_t$ is small, demonstrating the capability of the TopK-Chamfer Similarity in diminishing redundancy and reducing the influence of noise, thereby ensuring robustness in similarity calculation.

\section{Conclusion}
\label{sec:conclusion}

In this paper, we design a self-supervised framework for video retrieval, which features a video-oriented similarity measure to gather fine-grained features and a novel AP-based loss with reasonable gradients to correct mis-ranked instance pairs efficiently, filling the gap between the training objective and evaluation metric. 
Within the framework, we propose a hierarchical learning strategy to optimize AP on both video and frame levels, resulting in precise estimations of the AP loss and thus improving the accuracy of similarity learning.
Experimental results demonstrate that our framework often surpasses previous works in several benchmark datasets, making it a feasible solution for video retrieval tasks.

\begin{acks}
This work was supported in part by the National Key R\&D Program of China under Grant 2018AAA0102000, in part by National Natural Science Foundation of China: 62236008, U21B2038, U23B2051, 61931008,  62122075 and 61976202, in part by Youth Innovation Promotion Association CAS, in part by the Strategic Priority Research Program of the Chinese Academy of Sciences, Grant No. XDB0680000, in part by the Innovation Funding of ICT, CAS under Grant No.E000000. 
\end{acks}

\bibliographystyle{ACM-Reference-Format}
\bibliography{main}

\newpage
\clearpage

\appendix

\section*{Appendix}
\startcontents[appendices]
\printcontents[appendices]{l}{1}{\setcounter{tocdepth}{3}}

\section{Additional Illustration of Method}
\subsection{Derivation of AP Risk}
In a batch of videos $\bm{B} = \{\bm{V}_{i} \in \mathcal{X}\}_{i=1}^{N}$, recall that for a query video $\bm{V}_k$, the similarity scores of the relevant/irrelevant videos are denoted as $\bm{S}^{k+}$/$\bm{S}^{k-}$. 
For simplicity, let $d^k_{ji} = s_{kj} - s_{ki}$.
As mentioned in \cref{subsec:task_definition}, our goal is to maximize the AP score. This is achieved by minimizing the AP risk, which is derived as follows:
\begin{equation*}
\begin{aligned}
    AP^\downarrow_k(f) & =  1 - AP_k(f) \\
    & = 1 - \frac{1}{|\bm{S}^{k+}|} \sum_{s_{ki}\in \bm{S}^{k+}} \frac {\mathcal{R}(s_{ki}, \bm{S}^{k+})} {\mathcal{R}(s_{ki}, \bm{S}^{k+} \cup \bm{S}^{k-} )} \\ 
    & =  1 - \frac{1}{|\bm{S}^{k+}|} \sum_{s_{ki}\in \bm{S}^{k+}} \frac {
        1 + \sum_{s_{kj}\in \bm{S}^{k+}} \mathcal{H}(d^k_{ji})
    } {
        1 + \sum_{s_{kj}\in \bm{S}^{k+}\cup \bm{S}^{k-}} \mathcal{H}(d^k_{ji})
    } \\
    & = \frac{1}{|\bm{S}^{k+}|} \sum_{s_{ki}\in \bm{S}^{k+}} \frac {
        \sum_{s_{kj}\in \bm{S}^{k-}} \mathcal{H}(d^k_{ji})
    } {
        1 + \sum_{s_{kj}\in \bm{S}^{k+}\cup \bm{S}^{k-}} \mathcal{H}(d^k_{ji})
    } \\
    & = \frac{1}{|\bm{S}^{k+}|} \sum_{s_{ki}\in \bm{S}^{k+}} \frac {
        \sum_{s_{kj}\in \bm{S}^{k-}} \mathcal{H}(d^k_{ji})
    } {
        1 
        + \sum_{s_{kj}\in \bm{S}^{k+}} \mathcal{H}(d^k_{ji})
        + \sum_{s_{kj}\in \bm{S}^{k-}} \mathcal{H}(d^k_{ji})
    } \\
    & = \frac{1}{|\bm{S}^{k+}|} \sum_{s_{ki}\in \bm{S}^{k+}} 
    \frac {
       \left[ \sum_{s_{kj}\in \bm{S}^{k-}} \mathcal{H}(d^k_{ji}) \right] /
       \left[ 1 + \sum_{s_{kj}\in \bm{S}^{k+}} \mathcal{H}(d^k_{ji}) \right]
    } {
        1 + 
       \left[ \sum_{s_{kj}\in \bm{S}^{k-}} \mathcal{H}(d^k_{ji}) \right] /
       \left[ 1 + \sum_{s_{kj}\in \bm{S}^{k+}} \mathcal{H}(d^k_{ji}) \right]
    } \\
    & = \frac{1}{|\bm{S}^{k+}|} \sum_{s_{ki}\in \bm{S}^{k+}} h\left(\frac {
        \sum_{s_{kj}\in \bm{S}^{k-}} \mathcal{H}(d^k_{ji})
    } {
        1 + \sum_{s_{kj}\in \bm{S}^{k+}} \mathcal{H}(d^k_{ji})
    }\right),
\label{eq:ap_risk}
\end{aligned}
\end{equation*}

\noindent where $\mathcal{R}(s, \bm{S}) = 1+ \sum_{s' \in \bm{S}} 
\mathcal{H}(s' - s)$ is the descending ranking of $s$ in $\bm{S}$, $\mathcal{H}(\cdot)$ is the Heaviside function, $h(x) = \frac{x}{1 + x}$ is a monotonically increasing function.

We substitute the Heaviside function in the numerator with $\mathcal{R}^-(d^k_{ji};\delta)$ in eq.(10) and introduce an additional parameter $\rho$, which forms the following surrogate AP risk:

\begin{equation*}
    \widehat{AP}^\downarrow_k(f) = \frac{1}{|\bm{S}^{k+}|} \sum_{s_{ki}\in \bm{S}^{k+}} h\left(\frac {
        \sum_{s_{kj}\in \bm{S}^{k-}} \mathcal{R}^-(d^k_{ji};\delta)
    } {
        1 + \rho  \sum_{s_{kj}\in \bm{S}^{k+}} \mathcal{H}(d^k_{ji})
    }\right).
  \label{eq:QLAP_risk}
\end{equation*}


\subsection{Proofs of QuadLinear-AP's properties}
\label{subsec:supp_proof}
In this subsection, we provide proofs for several properties of QuadLinear-AP as outlined in \cref{subsec:video_oriented} of the main paper. Specifically, we focus on the proofs of $\mathcal{R}^-(x;\delta)$ since it determines these properties of QuadLinear-AP.

\subsubsection{Differentiability}
Note that it is unnecessary to replace $\mathcal{H}(\cdot)$ for the positive-positive pair since it only plays a role of weight for precisely measuring each term in eq. (8). Therefore, we only need to ensure the $\mathcal{R}^-(x;\delta)$ is differentiable, which is proved as follows.

First, the $\mathcal{R}^-(x;\delta)$ can be reformatted as:
\begin{equation*}
\begin{aligned}
\renewcommand{\arraystretch}{1.25}
\mathcal{R}^-(x;\delta) =
\left\{
  \begin{array}{ll}
    \frac{2}{\delta}x+1, & \text{ if } x \ge 0 . \\
    \frac{1}{\delta^2}x^2 + \frac{2}{\delta}x+1, & \text{ if } -\delta \le x < 0 . \\
    0, & \text{ if } x < -\delta .
  \end{array}
  \right.
  \label{eq:QL_R-}
\end{aligned}
\end{equation*}

Clearly, $\mathcal{R}^-(x;\delta)$ is differentiable on its three segments. Now, we only need to verify that it is differentiable at the points where $x=-\delta$ and $x=0$.

When $x=-\delta$, we have:
\begin{equation*}
\begin{aligned}
\frac{d\mathcal{R}^-(x^-;\delta)}{dx^-} & = \lim_{x^- \to -\delta} \frac{\mathcal{R}^-(x^-;\delta)-\mathcal{R}^-(-\delta;\delta)}{x^- - (-\delta)} = 0, \\
\frac{d\mathcal{R}^-(x^+;\delta)}{dx^+} & = \lim_{x^+ \to -\delta} \frac{\mathcal{R}^-(x^+;\delta)-\mathcal{R}^-(-\delta;\delta)}{x^+ - (-\delta)} = 0, \\
\frac{d\mathcal{R}^-(x^-;\delta)}{dx^-} & = \frac{d\mathcal{R}^-(x^+;\delta)}{dx^+} = \frac{d\mathcal{R}^-(x;\delta)}{dx}_{|x=-\delta} = 0.
  \label{eq:diff1}
\end{aligned}
\end{equation*}

When $x=0$, we have:
\begin{equation*}
\begin{aligned}
\frac{d\mathcal{R}^-(x^-;\delta)}{dx^-} & = \lim_{x^- \to 0} \frac{\mathcal{R}^-(x^-;\delta)-\mathcal{R}^-(0;\delta)}{x^- - 0} = \frac{2}{\delta}, \\
\frac{d\mathcal{R}^-(x^+;\delta)}{dx^+} & = \lim_{x^+ \to 0} \frac{\mathcal{R}^-(x^+;\delta)-\mathcal{R}^-(0;\delta)}{x^+ - 0} = \frac{2}{\delta}, \\
\frac{d\mathcal{R}^-(x^-;\delta)}{dx^-} & = \frac{d\mathcal{R}^-(x^+;\delta)}{dx^+} = \frac{d\mathcal{R}^-(x;\delta)}{dx}_{|x=0} = \frac{2}{\delta}.
  \label{eq:diff2}
\end{aligned}
\end{equation*}

Therefore, it is proven that $\mathcal{R}^-(x;\delta)$ is differentiable at each point, allowing backpropagation to be performed effectively during the optimization process to update model parameters. 
The derivative function of $\mathcal{R}^-(x;\delta)$ can be formulated as follows:

\begin{equation*}
\begin{aligned}
\renewcommand{\arraystretch}{1.25}
\frac{d\mathcal{R}^-(x;\delta)}{dx} =
\left\{
  \begin{array}{ll}
    \frac{2}{\delta}, & \text{ if } x \ge 0 . \\
    \frac{2}{\delta^2}x + \frac{2}{\delta}, & \text{ if } -\delta \le x < 0 . \\
    0, & \text{ if } x < -\delta .
  \end{array}
  \right.
  \label{eq:dQL_R-}
\end{aligned}
\end{equation*}

\subsubsection{Smoothness}
To prove the smoothness of $\mathcal{R}^-(x;\delta)$ is equivalent to proving that the derivation function of $\mathcal{R}^-(x;\delta)$ is continuous. This continuity is essential for ensuring stable gradient changes for efficient optimization and smooth convergence of the model. For the sake of presentation, let $\mathcal{D}^-(x;\delta) = \frac{d\mathcal{R}^-(x;\delta)}{dx}$.

Clearly, $\mathcal{D}^-(x;\delta)$ is continuous on its three segments, thus we only need to verify that it is continuous at the points where $x=-\delta$ and $x=0$, which is presented as follows:

\begin{equation*}
\begin{aligned}
\lim_{x^- \to -\delta}\mathcal{D}^-(x^-;\delta) &= \lim_{x^+ \to -\delta}\mathcal{D}^-(x^+;\delta) = \mathcal{D}^-(-\delta;\delta) = 0, \\
\lim_{x^- \to 0}\mathcal{D}^-(x^-;\delta) &= \lim_{x^+ \to 0}\mathcal{D}^-(x^+;\delta) = \mathcal{D}^-(0;\delta) = \frac{2}{\delta}.
  \label{eq:cont}
\end{aligned}
\end{equation*}

Therefore, it is proven that $\mathcal{R}^-(x;\delta)$ is smooth, and the derivative function of $\mathcal{R}^-(x;\delta)$ is continuous at each point.

\subsubsection{Convexity}
First, it is obvious that $\mathcal{R}^-(x;\delta)$ is convex on its three segments, thus we only need to verify three situations by proving $t\mathcal{R}^-(x_1;\delta) + \left(1-t\right)\mathcal{R}^-(x_2;\delta) - \mathcal{R}^-(tx_1+\left(1-t\right)x_2;\delta) \ge 0$ for the given $ 0 \le t \le 1$ and $x_1<x_2$.

\textbf{1)} When $-\delta \le x_1 < 0 \le x_2$, we have:
\begin{equation*}
\begin{aligned}
& t\mathcal{R}^-(x_1;\delta) + \left(1-t\right)\mathcal{R}^-(x_2;\delta) \\
& = t\left(\frac{1}{\delta^2}x_1^2+\frac{2}{\delta}x_1+1\right) + \left(1-t\right)\left(\frac{2}{\delta}x_2+1\right).
  \label{eq:conv1}
\end{aligned}
\end{equation*}

If $tx_1 + (1-t)x_2 \ge 0$ then:
\begin{equation*}
\begin{aligned}
& \mathcal{R}^-(tx_1+\left(1-t\right)x_2;\delta) = \frac{2}{\delta}\left[tx_1+\left(1-t\right)x_2\right]+1. \\
& t\mathcal{R}^-(x_1;\delta) + \left(1-t\right)\mathcal{R}^-(x_2;\delta) - \mathcal{R}^-(tx_1+\left(1-t\right)x_2;\delta) \\
&= \frac{t}{\delta^2}x_1^2 \\
&> 0.
  \label{eq:conv2}
\end{aligned}
\end{equation*}

If $ -\delta \le tx_1 + (1-t)x_2 < 0$ then:
\begin{equation*}
\begin{aligned}
& \mathcal{R}^-(tx_1+\left(1-t\right)x_2;\delta) = \left\{\frac{1}{\delta}\left[tx_1+\left(1-t\right)x_2\right]+1\right\}^2. \\
& t\mathcal{R}^-(x_1;\delta) + \left(1-t\right)\mathcal{R}^-(x_2;\delta) - \mathcal{R}^-(tx_1+\left(1-t\right)x_2;\delta) \\
&= \frac{t}{\delta^2}x_1^2 - \frac{1}{\delta^2}\left[tx_1+\left(1-t\right)x_2\right]^2 \\
&= \left[\frac{\sqrt{t}}{\delta}x_1 - \frac{t}{\delta}x_1 - \frac{1-t}{\delta}x_2 \right] \left[\frac{\sqrt{t}}{\delta}x_1 + \frac{t}{\delta}x_1 + \frac{1-t}{\delta}x_2 \right] \\
&> \left[\frac{\sqrt{t}}{\delta}x_1 - \frac{1}{\delta}x_2 \right] \left[\frac{\sqrt{t}}{\delta}x_1 + \frac{1}{\delta}x_1 \right]  \\
&> 0.
  \label{eq:conv3}
\end{aligned}
\end{equation*}

For the other two situations, \textit{i.e.}, $x_1 < -\delta < 0 \le x_2$ and $ x_1 < -\delta \le x_2 < 0$, the proof process is similar to the situation discussed above, and is therefore omitted for brevity.

In summary, $\mathcal{R}^-(x;\delta)$ is convex at each point, which facilitates finding the optimal solution while maintaining good convergence speed and stability.

\subsubsection{Non-strictly Monotonically Increasing}
First, it is obvious that $\mathcal{R}^-(x;\delta)$ is non-strictly monotonically increasing on its three segments, thus we only need to verify the following three situations:

\textbf{1)} When $-\delta \le x < 0$, for given $\varepsilon > 0$, if $x + \varepsilon \ge 0$ we have:
\begin{equation*}
\begin{aligned}
&\mathcal{R}^-(x+\varepsilon;\delta) - \mathcal{R}^-(x;\delta) \\
&= \frac{2}{\delta}\left(x+\varepsilon\right) + 1 - \left(\frac{1}{\delta^2}x^2 + \frac{2}{\delta}x+1\right) \\
&= \frac{1}{\delta}\left( 2\varepsilon-\frac{1}{\delta}x^2 \right) \\
&\ge \frac{1}{\delta}\left( -2x+\frac{1}{x} \cdot x^2 \right) \\
&> 0.
  \label{eq:incre1}
\end{aligned}
\end{equation*}

\textbf{2)} When $x < -\delta$, for given $\varepsilon > 0$, if $-\delta \le x + \varepsilon < 0$ we have:
\begin{equation*}
\begin{aligned}
& \mathcal{R}^-(x+\varepsilon;\delta) - \mathcal{R}^-(x;\delta) \\
&= \frac{1}{\delta^2}\left(x+\varepsilon\right)^2 + \frac{2}{\delta}\left(x+\varepsilon\right)+1 \\
&=\left[\frac{1}{\delta}\left( x+\varepsilon \right)\right]^2 \\
&> 0.
  \label{eq:incre2}
\end{aligned}
\end{equation*}

\textbf{3)} When $x < -\delta$, for given $\varepsilon > 0$, if $x + \varepsilon \ge 0$ we have:
\begin{equation*}
\begin{aligned}
\mathcal{R}^-(x+\varepsilon;\delta) - \mathcal{R}^-(x;\delta) &= \frac{2}{\delta}\left(x+\varepsilon\right) +1  > 0.
  \label{eq:incre3}
\end{aligned}
\end{equation*}

In summary, $\mathcal{R}^-(x;\delta)$ is non-strictly monotonically increasing, which can also be supported by the fig. 4c in the main paper.

\subsubsection{Upper Bound of Heaviside Function}
Here we prove $\mathcal{R}^-(x;\delta)$ is the upper bound of $\mathcal{H}(x)$, which is equivalent to prove the $\mathcal{R}^-(x;\delta)-\mathcal{H}(x) \ge 0$. Let $\mathcal{P}^-(x;\delta) = \mathcal{R}^-(x;\delta)-\mathcal{H}(x)$, we have:
\begin{equation*}
\begin{aligned}
\renewcommand{\arraystretch}{1.25}
\mathcal{P}^-(x;\delta) =
\left\{
  \begin{array}{ll}
    \frac{2}{\delta}x, & \text{ if } x \ge 0 . \\
    \frac{1}{\delta^2}x^2 + \frac{2}{\delta}x+1, & \text{ if } -\delta \le x < 0 . \\
    0, & \text{ if } x < -\delta .
  \end{array}
  \right.
  \label{eq:upper}
\end{aligned}
\end{equation*}
Obviously, $\mathcal{P}^-(x;\delta) \ge 0$, which illustrates the $\mathcal{R}^-(x;\delta)$ is the upper bound of $\mathcal{H}(x)$.

\subsection{Description of the Basic Loss}
\label{subsec:supp_basic_loss}

As outlined in \cref{subsec:hierarchical}, following previous methods on ranking optimization \cite{dai2024drauc,shao2024weighted}, we combine the AP losses with a basic loss $\mathcal{L}_{base}$, which comprises the InfoNCE loss~\cite{oord2018representation} and an SSHN loss~\cite{kordopatis2023self}.

The InfoNCE loss is widely used in self-supervised contrastive learning tasks due to its effectiveness and adaptability. For a query video $\bm{V}_k$, the InfoNCE loss is calculated by: 
\begin{equation*}
    \mathcal{L}_{NCE}^k = - \frac{1}{|\bm{S}^{k+}|} \sum_{s_{ki}\in \bm{S}^{k+}} log
    \frac {
        exp\left(s_{ki}/\tau\right)
    } {
        exp\left(s_{ki}/\tau\right) + \sum_{s_{kj}\in \bm{S}^{k-}} exp\left(s_{kj}/\tau\right)
    }.
\end{equation*}
Using the InfoNCE enables the model to support representation learning by distinguishing between positive and negative instances, thus promoting collaborative optimization between ranking and representation learning.

The SSHN loss promotes self-similarity towards 1 by compensating for the CNN block $\psi$, which tends to make $s_{kk}$ less than 1. Additionally, it performs hard negative mining by reducing the similarity of the most challenging negative instances, thus enhancing the distinction between similarities. The SSHN loss can be formulated as follows:
\begin{equation*}
    \mathcal{L}_{SSHN}^k = -log\left(s_{kk}\right)-log\left(\max_{s_{ki} \in \bm{S}^{k-}}\left(1-s_{ki}\right)\right).
\end{equation*}

Finally, we integrate these two losses as the following basic loss function, where $\lambda_s$ is hyperparameters to adjust the weights of the two losses.
\begin{equation}
    \mathcal{L}_{base} =  \frac{1}{N} \sum_{k=1}^N \left(\mathcal{L}_{NCE}^k+\lambda_s\mathcal{L}_{SSHN}^k\right).
\end{equation}

\subsection{Hierarchical Average Precision Optimization Algorithm}
\cref{alg:algo} outlines the process of our proposed hierarchical AP optimization method.

\label{sec:supp_algorithm}
\begin{algorithm}
\caption{Hierarchical Average Precision Optimization}
\begin{algorithmic}[1] 
\item[\textbf{Input:}] Training set $\bm{S}$, maximum iterations $L$, learning rate $\{\eta_l\}_{l=1}^L$, positive frame rate $r_t$, negative frame rate $r_b$.
\item[\textbf{Output:}] Model parameters $\bm{\Theta}_{L+1}$.
\State Initialize model parameters $\bm{\Theta}_1$.
\For {$l=1$ to $L$}
    \State Sample a batch of videos $\{\bm{V}_i\}_{i=1}^N$ form $\bm{S}$.
    \State Extract video embeddings $g(\bm{V}_i)$ and $g'(\bm{V}_i)$.
    \State Generate pseudo labels $\hat{Y}$ based on $r_t$ and $r_b$.
    \State Calculate similarities with function $f$ in ~\cref{eq:sim_score_fun}.
    \State Compute $\widehat{AP}^\downarrow_k(f)$ with ~\cref{eq:eq_QLAP_risk} to form $\mathcal{L}_{QLAP}^V$ and $\mathcal{L}_{QLAP}^F$.
    \State Compute the total loss $\mathcal{L}$ by ~\cref{eq:total_loss}.
    \State Update parameters: 
    $\bm{\Theta}_{l+1} = \bm{\Theta}_l - \eta_l  \nabla \mathcal{L}$.
\EndFor
\end{algorithmic}
\label{alg:algo}
\end{algorithm}

\section{Detailed Description of Experiments}
\label{sec:supp_detailed_experiments}
\subsection{Datasets}
The detailed description of the datasets used in our experiments is as follows:
\begin{itemize}
\item \textbf{VCDB}~\cite{jiang2014vcdb} is designed for the task of partial video copy detection. It contains a labeled core dataset denoted as VCDB ($\mathcal{C}$) and a large-scale unlabeled dataset with 100,000 distractor videos denoted as VCDB ($\mathcal{D}$). In our experiments, we only use the VCDB ($\mathcal{D}$) for self-supervised training.
\item \textbf{EVVE}~\cite{revaud2013event} is used as a benchmark video dataset for the task of event-based video retrieval. It includes 620 query videos and 2,373 database videos manually annotated into 13 event categories. Due to the absence of some videos, only 504 query videos and 1,906 database videos can be obtained. 
\item \textbf{SVD}~\cite{jiang2019svd} is designed for the task of near-duplicate video retrieval, containing 1,206 queries and 526,787 unlabelled videos in total. The dataset is organized into a training set and a test set. For evaluation in the experiments, we exclusively employ the test set, which includes 206 queries with 6,355 labeled video pairs and 526,787 unlabelled videos as distractors.
\item \textbf{FIVR-200K}~\cite{kordopatis2019fivr} is specifically designed for fine-grained incident video retrieval, comprising 100 queries and 225,960 database videos. It contains three distinct video retrieval subtasks: Duplicate Scene Video Retrieval (DSVR), Complementary Scene Video Retrieval (CSVR), and Incident Scene Video Retrieval (ISVR). Additionally, it also contains three distinct video detection subtasks: Duplicate Scene Video Detection (DSVD), Complementary Scene Video Detection (CSVD), and Incident Scene Video Detection (ISVD). 
\item \textbf{FIVR-5K}~\cite{kordopatis2023self}, a subset of FIVR-200K, which includes 50 queries and 5,000 database videos, containing the same subtasks as FIVR-200K. This dataset is also utilized in our experiments to facilitate swift comparative analysis.
\end{itemize}

Generally, we use the origin videos from VCDB ($\mathcal{D}$) to train our model and conduct evaluation on EVVE, SVD, FIVR-200K as well as FIVR-5K. Following the previous works~\cite{kordopatis2023self}, we use the extracted features of the evaluation datasets in our experiments. A summary of the descriptions for these datasets is presented in ~\cref{tab:dataset}.

\begin{table*}[htbp]
  \centering
  \caption{Summary of the descriptions for the VCDB, EVVE, SVD, FIVR-200K, and FIVR-5K datasets.}
  \resizebox{0.96\textwidth}{!}{%
    \begin{tabular}{cc|lcc}
    \toprule
    \multicolumn{2}{c|}{\textbf{Dataset}} & \multicolumn{1}{c}{\textbf{Video Task}} & \textbf{\# of Query Videos} & \textbf{\# of Database Videos} \\
    \midrule
    \multicolumn{2}{c|}{VCDB} & Partial Video Copy Detection & 528   & 100,000 \\
    \multicolumn{2}{c|}{EVVE} & Event-based Video Retrieval & 620   & 2,373 \\
    \multicolumn{2}{c|}{SVD} & Near-duplicate Video Retrieval & 206   & 526,787 \\
    FIVR-200K & DSVR / DSVD & Duplicate Scene Video Retrieval / Detection & 200   & 225,960 \\
    FIVR-200K & CSVR / CSVD & Complementary Scene Video Retrieval / Detection & 200   & 225,960 \\
    FIVR-200K & ISVR / ISVD & Incident Scene Video Retrieval / Detection & 200   & 225,960 \\
    FIVR-5K & DSVR / DSVD & Duplicate Scene Video Retrieval / Detection & 50   & 5,000 \\
    FIVR-5K & CSVR / CSVD & Complementary Scene Video Retrieval / Detection & 50   & 5,000 \\
    FIVR-5K& ISVR / ISVD & Incident Scene Video Retrieval / Detection & 50   & 5,000 \\
    \bottomrule
    \end{tabular}%
    }
  \label{tab:dataset}%
\end{table*}%

\subsection{Evaluation Metrics}

\textbf{Mean Average Precision}
Mean Average Precision (\textbf{mAP}), also known as macro Average Precision~\cite{perronnin2009family}, serves as the primary metric to evaluate the overall performance of retrieval tasks. Specifically, AP computes the average ranking of positive instances in the retrieval set for a particular query, while mAP calculates the mean of these AP values across all queries. The definition of mAP is given in \cref{eq:mAP}, where $n_{j}$ denotes the number of positive instances for a particular query, $r_{i}$ represents the ranking of the i-th retrieved positive instance in the retrieval set, and $|\bm{Q}|$ is the number of query instances.
\begin{equation}
mAP = \frac{1}{|\bm{Q}|} \sum_{j=1}^{|\bm{Q}|} \frac{1}{n_j} \sum_{i=1}^{n_j} \frac{i}{r_i}
  \label{eq:mAP}
\end{equation}

\textbf{Micro Average Precision}
Micro Average Precision (\textbf{$\bm{\mu}$AP}) is a metric employed in prior research~\cite{law2007video, pizzi2022self, kordopatis2023self} to evaluate the performance of detection tasks. In contrast to mAP, $\mu$AP considers the joint distribution of similarities across all queries by calculating the AP across all queries simultaneously, which reflects the model's capability to consistently apply a uniform threshold across various queries to detect relevant instances. $\mu$AP is computed as outlined in \cref{eq:uAP}, where $|\bm{R}|$ is the number of all reference instances, $p(i)$ represents the precision of i-th instance and $\Delta r(i)$ denotes the difference of recall between i-th and its adjacent instance in the sorted list according to similarity scores.
\begin{equation}
\mu AP = \sum_{i=1}^{|\bm{R}|}p(i) \Delta r(i)
  \label{eq:uAP}
\end{equation}



\subsection{Implementation Details}
\label{subsec:implementation_details}
In this subsection, we provide additional descriptions of implementation details including the data processing, experiment configuration, and hyperparameter settings. 

\textbf{Data processing}
We adopt a self-supervised learning approach as introduced in ~\cite{kordopatis2023self}, where videos in a batch are subjected to weak and strong augmentations to simulate common video copy transformations in actual situations.
The weak augmentation function set $\bm{A}_w$ includes traditional geometric transformations such as random cropping and horizontal flipping, applied to the frames of the entire video. 
The strong augmentation function set $\bm{A}_s$, on the other hand, involves more complex transformations: 
\textbf{1)} \textit{Global transformations} apply different geometric and optical image transformations on all frames consistently by RandAugment~\cite{cubuk2020randaugment}; 
\textbf{2)} \textit{Frame transformations} encompass overlaying emojis and text on randomly selected frames and applying blur to frames~\cite{pizzi2022self}; 
\textbf{3)} \textit{Temporal transformations}, including fast forward, slow motion, reverse play, frame pause, and sub-clip shuffle/dropout~\cite{kordopatis2019visil,kordopatis2023self}, are utilized to create intense temporal manipulations; 
\textbf{4)} \textit{Video mix-up transformation}, down-scales a video and embeds it within another video~\cite{kordopatis2023self}.

\textbf{Experiment configuration}
For the training video data, following the previous work ~\cite{shao2021temporal,kordopatis2019visil,kordopatis2022dns}, we first extract one frame per second for each video.
Subsequently, we resize the frames to 256 pixels and crop them to 224 pixels, then randomly select 28 consecutive frames to constitute a video clip.
For the backbone feature extractor $g(\cdot)$, following previous literature~\cite{kordopatis2019visil,kordopatis2022dns,kordopatis2023self}, we adopt ResNet50~\cite{he2016deep} pretrained on ImageNet~\cite{deng2009imagenet}. 
The backbone feature extractor $g(\cdot)$ performs the mapping $g: \mathbb{R}^{T \times H \times W \times C} \rightarrow \mathbb{R}^{T \times R \times D}$, where $T=28, H=224, W=224, C=3, R=9, D=512$.
For the feature extractor $g'(\cdot)$ in the pseudo label generator, we utilize 
DINO~\cite{caron2021emerging} pretrained ViT-small~\cite{dosovitskiy2020image} with a patch size of 16. 
The feature extractor $g'(\cdot)$ performs the mapping $g': \mathbb{R}^{T \times H \times W \times C} \rightarrow \mathbb{R}^{T \times D'}$, where $T=28, H=224, W=224, C=3, D'=384$. 

\textbf{Hyperparameter settings}
Our model is trained for 30,000 iterations with a batch size of 64. We use AdamW~\cite{loshchilov2018decoupled} with the Cosine Annealing scheduler for parameters optimization. The learning rate is set to $4 \times 10^{-5}$ with a warm-up period \cite{loshchilov2016sgdr} of 1,000 iterations, and weight decay is set to $1 \times 10^{-2}$. 
For the hyperparameters concerning QuadLinear-AP, we choose $\delta_v=0.05$, $\rho_v=0.10$ for $\mathcal{L}^V_{QLAP}$, and $\delta_f=0.05$, $\rho_f=5.00$ for $\mathcal{L}^F_{QLAP}$.
The weights of $\mathcal{L}^V_{QLAP}$ and $\mathcal{L}^F_{QLAP}$ are selected as $\lambda_v=4$ and $\lambda_f=6$.
The top and bottom rates for dividing positive and negative frame instances in the pseudo label generator are set to $r_t=0.35$ and $r_b=0.35$. The top-k rates of TopK-Chamfer Similarity within spacial and temporal correlation aggregation are set to $k_s=0.10$ and $k_t=0.03$, respectively.

Generally, the settings and hyperparameters for our HAP-VR framework within the training process are summarized in ~\cref{tab:hyperparams}. All experiments in this work are conducted with Pytorch~\cite{paszke2019pytorch} library on a Linux machine equipped with an Intel Gold 6230R CPU and two NVIDIA 3090 GPUs.

\begin{table}[htbp]
  \centering
  \small
  \caption{The settings and hyperparameters for our HAP-VR framework within the training process.}
  \resizebox{0.42\textwidth}{!}{%
    \begin{tabular}{lcc}
    \toprule
    \textbf{Hyperparameter} & \textbf{Notation} & \textbf{Value} \\
    \midrule
    \multicolumn{3}{c}{Training process} \\
    \midrule
    Iterations &    /   & 30,000 \\
    Warm-up iterations &   /    & 1,000 \\
    Batch size &    /   & 64 \\
    Learning rate &   /    & $4 \times 10^{-5}$ \\
    Optimizer &   /    & AdamW \\
    Learning rate scheduler &   /    & Cosine \\
    Weight decay &    /   & $1 \times 10^{-2}$ \\
    \midrule
    \multicolumn{3}{c}{Backbone feature extractor} \\
    \midrule
    \# of frames in a clip &    $T$   & 28 \\
    Frame size &    $H$,$W$   & 224 \\
    \# of ResNet50 feature patch &    $R$   & 9 \\
    \# of ResNet50 feature dim. &    $D$   & 512 \\
    \midrule
    \multicolumn{3}{c}{Pseudo label generator} \\
    \midrule
    \# of frames in a clip &    $T$   & 28 \\
    Frame size &    $H$,$W$   & 224 \\
    \# of ViT-small feature dim. &    $D'$   & 384 \\
    ViT-small patch size &   /   & 16 \\
    Top frame rate &   $r_t$    & 0.35 \\
    Bottom frame rate &    $r_b$   & 0.35 \\
    \midrule
    \multicolumn{3}{c}{QuadLinear-AP} \\
    \midrule
    Video-level Pos-neg margin  &    $\delta_v$   & 0.05 \\
    Video-level Pos-pos weight &   $\rho_v$    & 0.10 \\
    Video-level AP loss weight &   $\lambda_v$    & 4.00 \\
    Frame-level Pos-neg margin  &    $\delta_f$   & 0.05 \\
    Frame-level Pos-pos weight &   $\rho_f$    & 5.00 \\
    Frame-level AP loss weight &   $\lambda_f$    & 6.00 \\
    \midrule
    \multicolumn{3}{c}{TopK-Chamfer Similarity} \\
    \midrule
    Spacial top-k rate &   $k_s$    & 0.10 \\
    Temporal top-k rate &   $k_t$   & 0.03 \\
    \bottomrule
    \end{tabular}%
    }
  \label{tab:hyperparams}%
\end{table}%

\subsection{Additional Ablation Study}
In this section, we explore the impact of hyperparameters in our framework on performance. Except for the specific hyperparameters being investigated, we maintain consistency in all other experimental settings to ensure a fair comparison.


\textbf{Impact of $\delta_v$ and $\delta_f$} 
The results of our model trained with different $\delta_v$ and $\delta_f$ are presented in \cref{tab:ablation_for_video_delta} and \cref{tab:ablation_for_frame_delta}, respectively. The performance decreases for both hyperparameters when set above or below 0.05. 
This highlights the importance of selecting the appropriate $\delta_v$ and $\delta_f$ values to effectively balance the margin for correctly ranked positive-negative pairs and the penalty for incorrectly ranked positive-negative pairs.

\textbf{Impact of $\rho_v$ and $\rho_f$} 
The results of our model trained with different $\rho_v$ and $\rho_f$ are shown in \cref{tab:ablation_for_video_rho} and \cref{tab:ablation_for_frame_rho}, respectively.
For $\mathcal{L}^V_{QLAP}$, setting $\rho_v$ as a small value such as 0.2 provides optimal benefits as it more effectively adjusts the weight of positive-positive pairs and thus achieves a trade-off with positive-negative pairs. For $\mathcal{L}^F_{QLAP}$, which deals with more ambiguous inter-frame correlations, tuning $\rho_f$ within the range of 0.2 to 5 allows the model to better adapt to the varying instance distributions across different subtasks.

\textbf{Impact of $\lambda_v$ and $\lambda_f$}
In \cref{tab:ablation_for_lambda}, we report the results of our model trained with various values of $\lambda_f$ while keeping $\lambda_v$ fixed at 4 to simplify the comparative analysis. It can be observed that increasing $\lambda_f$ beyond $\lambda_v$ leads to an obvious performance gain. This is expected as more challenging frame-level similarities require greater weight for effective optimization. Furthermore, finding a balance between $\lambda_f$ and $\lambda_v$ with the weight of $\mathcal{L}_{base}$ can jointly promote ranking and representation learning, thereby enhancing the overall performance of the model.

\textbf{Impact of $r_t$ and $r_b$}
In \cref{tab:ablation_for_top_bottom_rate}, we report the results of our model trained with various combinations of $r_t$ and $r_b$.
When $r_t$ and $r_b$ are equal, setting higher values leads to similar frames being forcibly divided as positive and negative instances, thereby decreasing the model's discriminating ability. Conversely, setting lower values may cause the model to focus only on easier instances, thus resulting in insufficient learning and optimization.
When $r_t$ and $r_b$ are different, the performance tends to decrease due to the uneven distribution of positive and negative instances increasing the complexity of similarity learning.

\begin{table}[htbp]
  \centering
    \small
     \caption{Results on FIVR-5K in video retrieval and detection tasks with mAP (\%) and $\bm{\mu}$AP (\%) for $\delta_v$ within $\mathcal{L}^V_{QLAP}$. The first and second best results are marked with \textbf{bold} and \underline{underline}.}
    \resizebox{0.42\textwidth}{!}{%
    \setlength{\extrarowheight}{1pt}
    \begin{tabular}{c|ccc|ccc}
    \toprule
    \multirow{2}[2]{*}{$\delta_v$} & \multicolumn{3}{c|}{\textbf{Retrieval}} & \multicolumn{3}{c}{\textbf{Detection}} \\
\cmidrule{2-7} & DSVR  & CSVR  & ISVR  & DSVD  & CSVD  & ISVD \\
    \midrule
    0.01   & 87.32 & 86.66 & 80.27 & 73.53 & 72.15 & 63.58 \\
    0.05   & \textbf{88.86} & \textbf{87.79} & 80.34 & \textbf{78.15} & \textbf{76.29} & \textbf{65.88} \\
    0.10   & \underline{88.52} & \underline{87.38} & \underline{80.37} & \underline{76.94} & \underline{75.40} & \underline{65.60} \\
    0.15   & 87.34 & 86.62 & \textbf{80.42} & 73.27 & 72.37 & 64.11 \\
    \bottomrule
    \end{tabular}%
    }
  \label{tab:ablation_for_video_delta}%
\end{table}%

\begin{table}[htbp]
  \centering
     \caption{Results on FIVR-5K in video retrieval and detection tasks with mAP (\%) and $\bm{\mu}$AP (\%) for $\delta_f$ within $\mathcal{L}^F_{QLAP}$. The first and second best results are marked with \textbf{bold} and \underline{underline}.}
    \resizebox{0.42\textwidth}{!}{%
    \setlength{\extrarowheight}{1pt}
    \begin{tabular}{c|ccc|ccc}
    \toprule
    \multirow{2}[2]{*}{$\delta_f$} & \multicolumn{3}{c|}{\textbf{Retrieval}} & \multicolumn{3}{c}{\textbf{Detection}} \\
\cmidrule{2-7} & DSVR  & CSVR  & ISVR  & DSVD  & CSVD  & ISVD \\
    \midrule
    0.01   & \underline{90.30} & \textbf{89.19 }& \textbf{81.52} & 81.17 & 78.37 & \textbf{69.98} \\
    0.05   & \textbf{90.37} & \underline{89.11} & \underline{80.94} & \textbf{83.62} & \textbf{80.47} & \underline{69.30} \\
    0.10 & 89.85 & 88.52 & 79.93 & \underline{82.29} & \underline{79.57} & 67.15 \\
    0.15 & 89.55 & 88.05 & 79.37 & \underline{82.29} & 79.40 & 65.78 \\
    \bottomrule
    \end{tabular}%
    }
  \label{tab:ablation_for_frame_delta}%
\end{table}%

\begin{table}[htbp]
  \centering
    \small
     \caption{Results on FIVR-5K in video retrieval and detection tasks with mAP (\%) and $\bm{\mu}$AP (\%) for $\rho_v$ within $\mathcal{L}^V_{QLAP}$. The first and second best results are marked with \textbf{bold} and \underline{underline}.}
    \resizebox{0.43\textwidth}{!}{%
    \setlength{\extrarowheight}{1pt}
    \begin{tabular}{c|ccc|ccc}
    \toprule
    \multirow{2}[2]{*}{$\rho_v$} & \multicolumn{3}{c|}{\textbf{Retrieval}} & \multicolumn{3}{c}{\textbf{Detection}} \\
\cmidrule{2-7} & DSVR  & CSVR  & ISVR  & DSVD  & CSVD  & ISVD \\
    \midrule
    0.02   & 89.52 & 88.45 & 80.84 & 78.34 & 76.39 & 65.98 \\
    0.2   & \textbf{90.28} & \textbf{89.10} & \textbf{81.09} & \textbf{81.39} & \textbf{78.64} & \underline{68.79} \\
    1.0   & \underline{89.89} & \underline{88.81} & \underline{81.03} & \underline{80.41} & \textbf{78.64} & \textbf{69.41} \\
    5.0   & 89.51 & 88.31 & 80.77 & 80.00 & \underline{77.59} & 67.20 \\
    50   & 89.37 & 88.16 & 80.89 & 79.92 & 77.19 & 66.91 \\
    \bottomrule
    \end{tabular}%
    }
  \label{tab:ablation_for_video_rho}%
\end{table}%

\begin{table}[htbp]
  \centering
    \small
     \caption{Results on FIVR-5K in video retrieval and detection tasks with mAP (\%) and $\bm{\mu}$AP (\%) for $\rho_f$ within $\mathcal{L}^F_{QLAP}$. The first and second best results are marked with \textbf{bold} and \underline{underline}.}
    \resizebox{0.43\textwidth}{!}{%
    \setlength{\extrarowheight}{1pt}
    \begin{tabular}{c|ccc|ccc}
    \toprule
    \multirow{2}[2]{*}{$\rho_f$} & \multicolumn{3}{c|}{\textbf{Retrieval}} & \multicolumn{3}{c}{\textbf{Detection}} \\
\cmidrule{2-7} & DSVR  & CSVR  & ISVR  & DSVD  & CSVD  & ISVD \\
    \midrule
    0.02  & 89.71 & 88.01 & 79.44 & 82.73 & 78.98 & 65.80 \\
    0.2   & \underline{90.21} & 89.01 & 80.92 & \textbf{83.81} & \textbf{80.59} & 69.13 \\
    1.0   & \textbf{90.37} & \textbf{89.11} & 80.94 & \underline{83.62} & \underline{80.47} & 69.30  \\
    5.0   & 90.17 & \underline{89.07} & \textbf{81.29} & 82.93 & 80.18 & \textbf{70.99 }\\
    50  & 88.77 & 87.74 & \underline{81.18} & 79.76 & 78.46 & \underline{70.73} \\
    \bottomrule
    \end{tabular}%
    }
  \label{tab:ablation_for_frame_rho}%
\end{table}%

\begin{table}[htbp]
  \centering
     \caption{Results on FIVR-5K in video retrieval and detection tasks with mAP (\%) and $\bm{\mu}$AP (\%) for the weight of $\mathcal{L}^F_{QLAP}$, \textit{i.e.}, $\lambda_f$. The first and second best results are marked with \textbf{bold} and \underline{underline}.}
    \resizebox{0.42\textwidth}{!}{%
    \setlength{\extrarowheight}{1pt}
    \begin{tabular}{c|ccc|ccc}
    \toprule
    \multirow{2}[2]{*}{$\lambda_f$} & \multicolumn{3}{c|}{\textbf{Retrieval}} & \multicolumn{3}{c}{\textbf{Detection}} \\
\cmidrule{2-7} & DSVR  & CSVR  & ISVR  & DSVD  & CSVD  & ISVD \\
    \midrule
    2   & 89.73 & 88.50 & 80.42 & 80.45 & 77.89 & 66.32 \\
    4   & \underline{90.21} & \underline{89.08} & \underline{81.33} & \underline{83.15} & \underline{80.69} & \underline{70.61} \\
    6   & \textbf{90.26} & \textbf{89.18} & \textbf{81.47} & 83.05 & 80.11 & 70.32 \\
    8   & 90.07 & 88.91 & 81.30 & \textbf{84.20} & \textbf{80.97} & \textbf{70.63} \\
    \bottomrule
    \end{tabular}%
    }
  \label{tab:ablation_for_lambda}%
\end{table}%

\begin{table}[htbp]
  \centering
    \small
     \caption{Results on FIVR-5K in video retrieval and detection tasks with mAP (\%) and $\bm{\mu}$AP (\%) for $r_t$ and $r_b$. The first and second best results are marked with \textbf{bold} and \underline{underline}.}
    \resizebox{0.47\textwidth}{!}{%
    \setlength{\extrarowheight}{1pt}
    \begin{tabular}{cc|ccc|ccc}
    \toprule
    \multirow{2}[2]{*}{$r_t$} & \multirow{2}[2]{*}{$r_b$} & \multicolumn{3}{c|}{\textbf{Retrieval}} & \multicolumn{3}{c}{\textbf{Detection}} \\
\cmidrule{3-8} &  & DSVR  & CSVR  & ISVR  & DSVD  & CSVD  & ISVD \\
    \midrule
    0.30 & 0.30  & \underline{90.17} & 88.92 & 80.90 & \underline{83.08} & 80.10 & 69.02 \\
    0.35 & 0.35  & \textbf{90.21} & \textbf{89.08} & \textbf{81.33} & \textbf{83.15} & \textbf{80.69} & \underline{70.61} \\
    0.40 & 0.40  & \underline{90.17} & \underline{89.07} & \underline{81.29} & 82.93 & \underline{80.18} & \textbf{70.99}  \\
    0.45 & 0.45  & 89.63 & 88.37 & 80.89 & 82.28 & 79.38 & 70.18 \\
    0.30 & 0.40  & 89.73 & 88.62 & 81.18 & 82.67 & 79.68 & 69.18 \\
    0.40 & 0.30  & 89.82 & 88.58 & 80.62 & 82.28 & 79.33 & 69.18 \\
    \bottomrule
    \end{tabular}%
    }
  \label{tab:ablation_for_top_bottom_rate}%
\end{table}%

\begin{figure*}
  \centering
    \includegraphics[width=0.98\linewidth]{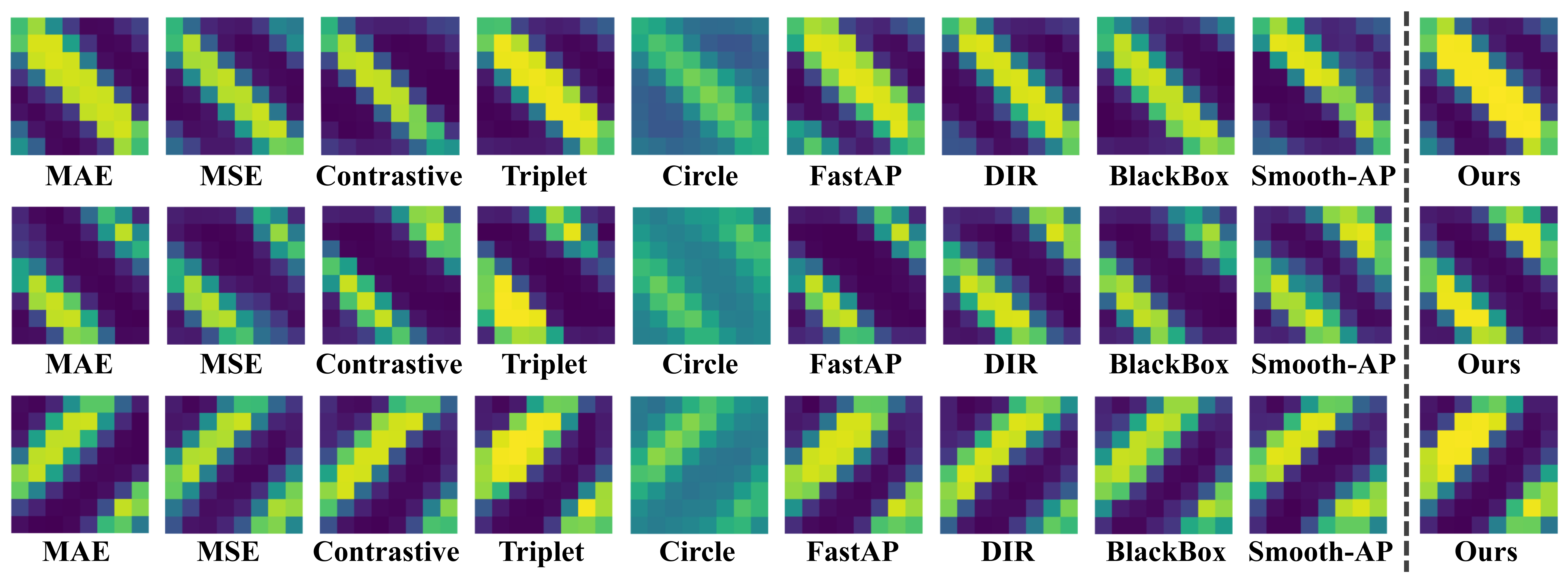}
    \caption{Heatmaps of frame-level similarity matrices generated by various losses. In contrast, our QuadLinear-AP distinguishes between relevant and irrelevant instances more clearly. A brighter color indicates a higher similarity score.}
    \label{fig:heatmaps}
\end{figure*}

\section{Visualization}

In this section, we provide examples to compare frame-level similarity matrices under different losses through visualization for intuitive analysis. The experiment settings remain consistent with those described in \cref{subsec:ablation_study} of the main paper. By analyzing these heatmaps, we can make the following observations:

\textbf{1)} Circle loss struggles to distinguish instances clearly even after parameter adjustments, likely due to its sensitivity of data distribution making it perform poorly in challenging video data with an imbalanced distribution.
\textbf{2)} Triplet loss is prone to become overconfident, which may lead to more irrelevant instances being predicted as relevant, thus increasing the risk of overfitting.
\textbf{3)} While other loss functions can discriminate between instances, there is still room for improvement in their performance.
\textbf{4)} Compared to other competitors, our proposed QuadLinear-AP generates higher similarity scores for relevant pairs of frames while lower scores for the irrelevant ones, thereby creating a clearer fine-grained distinction between the pairs compared to other competitors.
provides a clearer distinction between relevant and irrelevant instances, making it effective for video retrieval tasks.









\end{document}